\documentclass[journal]{IEEEtran}
\usepackage{graphicx,subfigure,amsmath,enumitem,algorithm,algorithmic,amssymb,tabu,multirow}

\newcommand{\tabincell}[2]{\begin{tabular}{@{}#1@{}}#2\end{tabular}}

\ifCLASSINFOpdf
\else
\fi
\begin{document}
%
\title{Deep Multi-Center Learning for Face Alignment}
%
%
%

\author{Zhiwen Shao$^{1*}$, Hengliang Zhu$^1$, Xin Tan$^1$, Yangyang Hao$^1$, and Lizhuang Ma$^{2,1*}$
\\$^1$Department of Computer Science and Engineering, Shanghai Jiao Tong University, China
\\$^2$School of Computer Science and Software Engineering, East China Normal University, China
\\\{shaozhiwen, hengliang\_zhu, tanxin2017, haoyangyang2014\}@sjtu.edu.cn, ma-lz@cs.sjtu.edu.cn
\thanks{$^*$ Corresponding author.}}
\maketitle

\begin{abstract}
Facial landmarks are highly correlated with each other since a certain landmark can be estimated by its neighboring landmarks. Most of the existing deep learning methods only use one fully-connected layer called shape prediction layer to estimate the locations of facial landmarks. In this paper, we propose a novel deep learning framework named Multi-Center Learning with multiple shape prediction layers for face alignment. In particular, each shape prediction layer emphasizes on the detection of a certain cluster of semantically relevant landmarks respectively. Challenging landmarks are focused firstly, and each cluster of landmarks is further optimized respectively. Moreover, to reduce the model complexity, we propose a model assembling method to integrate multiple shape prediction layers into one shape prediction layer. Extensive experiments demonstrate that our method is effective for handling complex occlusions and appearance variations with real-time performance. The code for our method is available at https://github.com/ZhiwenShao/MCNet-Extension.
\end{abstract}


%
\IEEEpeerreviewmaketitle

\section{Introduction}
\label{sec:intro}

Face alignment refers to detecting facial landmarks such as eye centers, nose tip, and mouth corners. It is the preprocessor stage of many face analysis tasks like face animation \cite{cao2014displaced}, face beautification \cite{fan2016efficient}, and face recognition \cite{leng2016cascade}. A robust and accurate face alignment is still challenging in unconstrained scenarios, owing to severe occlusions and large appearance variations. Most conventional methods \cite{cao2012face,xiong2013supervised,burgos2013robust,ren2014face} only use low-level handcrafted features and are not based on the prevailing deep neural networks, which limits their capacity to represent highly complex faces.

Recently, several methods use deep networks to estimate shapes from input faces. Sun et al. \cite{sun2013deep}, Zhou et al. \cite{zhou2013extensive}, and Zhang et al. \cite{zhang2014coarse} employed cascaded deep networks to refine predicted shapes successively. Due to the use of multiple networks, these methods have high model complexity with complicated training processes. Taking this into account, Zhang et al. \cite{zhang2014facial,zhang2015learning} proposed a Tasks-Constrained Deep Convolutional Network (TCDCN), which uses only one deep network with excellent performance. However, it needs extra labels of facial attributes for training samples, which limits its universality.

\begin{figure}[!htb]
  \centering
  \subfigure[Chin is occluded.]{
    \label{fig:unconstrained:a} 
    \includegraphics[width=1.2in]{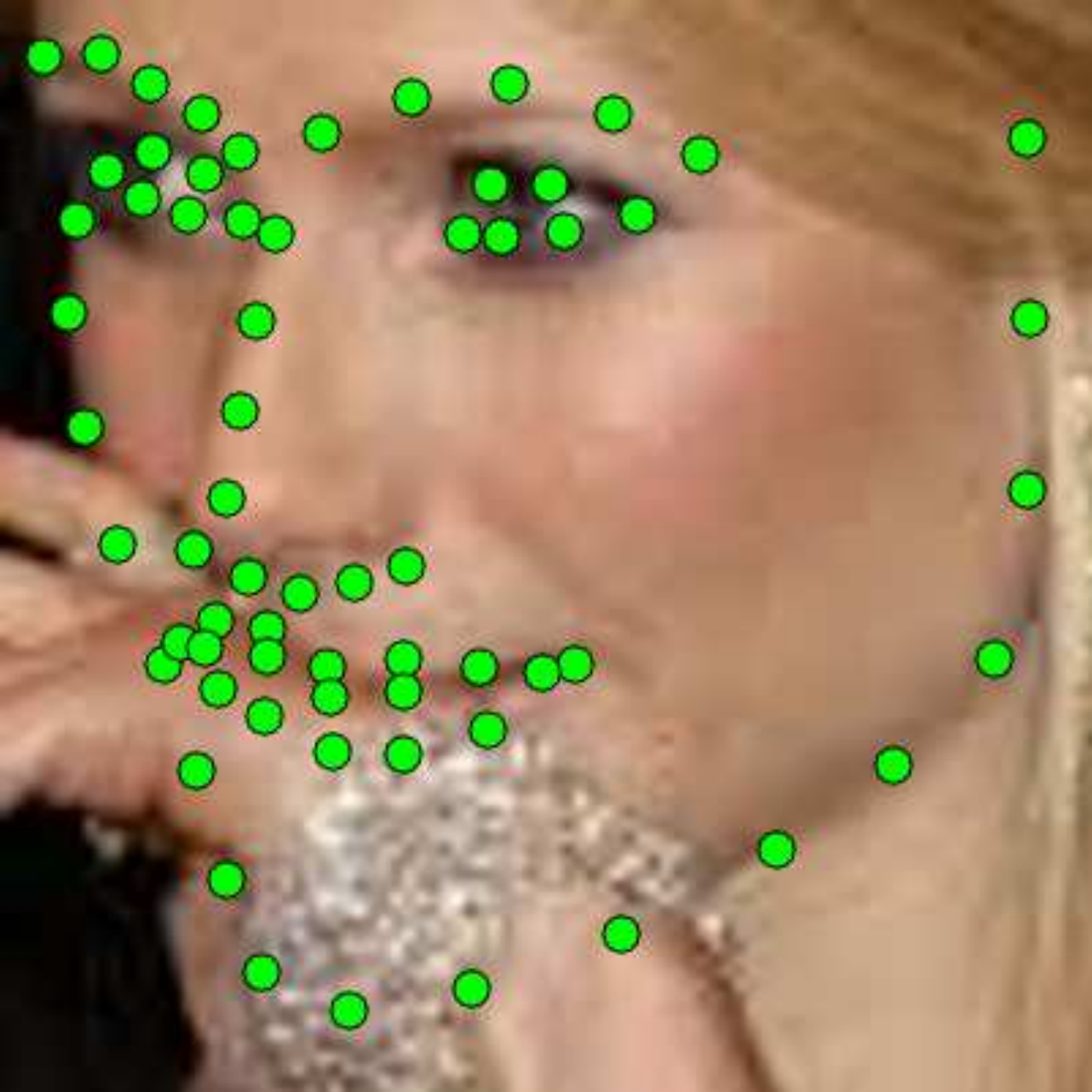}}
  \hspace{0.2in}
  \subfigure[Right contour is invisible.]{
    \label{fig:unconstrained:b} 
    \includegraphics[width=1.2in]{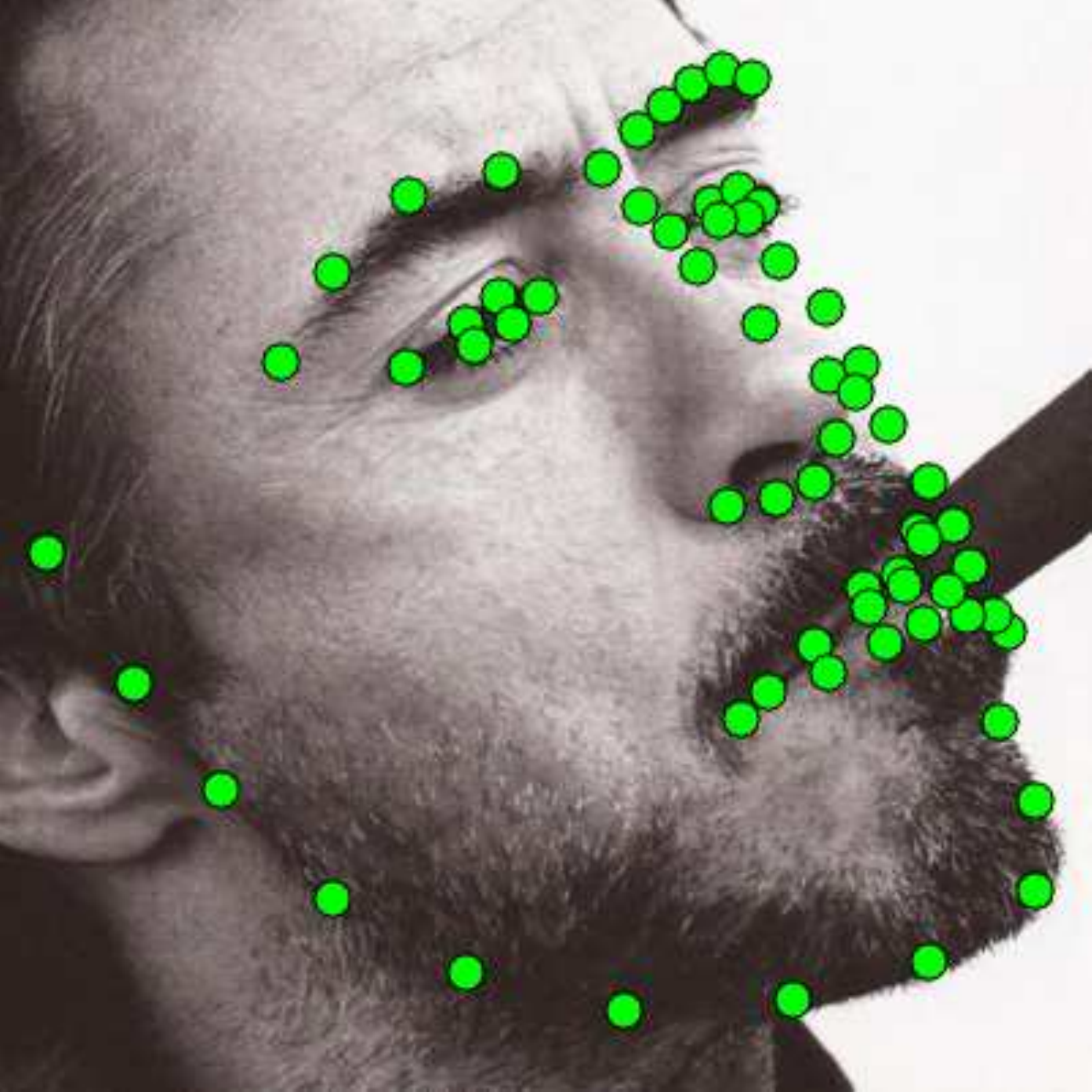}}
  \caption{Examples of unconstrained face images with partial occlusions and large pose.}
  \label{fig:unconstrained} 
\end{figure}

Each facial landmark is not isolated but highly correlated with adjacent landmarks. As shown in Fig. \ref{fig:unconstrained:a}, facial landmarks along the chin are all occluded, and landmarks around the mouth are partially occluded. Fig. \ref{fig:unconstrained:b} shows that landmarks on the right side of face are almost invisible. Therefore, landmarks in the same local face region have similar properties including occlusion and visibility. It is observed that the nose can be localized roughly with the locations of eyes and mouth. There are also structural correlations among different facial parts. Motivated by this fact, facial landmarks are divided into several clusters based on their semantic relevance.

In this work\footnote{This is an extended version of \cite{shao2017learning} with two improvements. The shape prediction layer is replaced from the fully-connected layer to the Global Average Pooling layer \cite{lin2013network}, which has a stronger feature learning ability. To exploit the correlations among landmarks more completely, challenging landmarks are focused firstly before each cluster of landmarks is respectively optimized.}, we propose a novel deep learning framework named Multi-Center Learning (MCL) to exploit the strong correlations among landmarks. In particular, our network uses multiple shape prediction layers to predict the locations of landmarks, and each shape prediction layer emphasizes on the detection of a certain cluster of landmarks respectively. By weighting the loss of each landmark, challenging landmarks are focused firstly, and each cluster of landmarks is further optimized respectively. Moreover, to decrease the model complexity, we propose a model assembling method to integrate multiple shape prediction layers into one shape prediction layer. The entire framework reinforces the learning process of each landmark with a low model complexity.

The main contributions of this study can be summarized as follows:
\begin{itemize}
    \item We propose a novel multi-center learning framework for exploiting the strong correlations among landmarks.
    \item We propose a model assembling method which ensures a low model complexity.
    \item Extensive experiments demonstrate that our method is effective for handling complex occlusions and appearance variations with real-time performance.
\end{itemize}

The remainder of this paper is structured as below. We discuss related works in the next section. In Section \ref{sec:mcnet}, we illuminate the structure of our network and the learning algorithm. Extensive experiments are carried out in Section \ref{sec:experi}. Section \ref{sec:concl} concludes this work.

\section{Related Work}
\label{sec:related}

We review researches from three aspects related to our method: conventional face alignment, unconstrained face alignment, face alignment via deep learning.

\subsection{Conventional Face Alignment}

Conventional face alignment methods can be classified as two categories: template fitting and regression-based.

Template fitting methods match faces by constructing shape templates. Cootes et al. \cite{cootes2001active} proposed a typical template fitting method named Active Appearance Model (AAM), which minimizes the texture residual to estimate the shape. Asthana et al. \cite{asthana2013robust} used regression techniques to learn functions from response maps to shapes, in which the response map has stronger robustness and generalization ability than texture based features of AAM. Pedersoli et al. \cite{pedersoli2014using} developed the mixture of trees of parts method by extending the mixtures from trees to graphs, and learned a deformable detector to align its parts to faces. However, these templates are not complete enough to cover complex variations, which are difficult to be generalized to unseen faces.

Regression-based methods predict the locations of facial landmarks by learning a regression function from face features to shapes. Cao et al. \cite{cao2012face} proposed an Explicit Shape Regression (ESR) method to predict the shape increment with pixel-difference features. Xiong et al. \cite{xiong2013supervised} proposed a Supervised Descent Method (SDM) to detect landmarks by solving the nonlinear least squares problem, with Scale-Invariant Feature Transform (SIFT) \cite{lowe2004distinctive} features and linear regressors being applied. Ren et al. \cite{ren2014face} used a locality principle to extract a set of Local Binary Features (LBF), in which a linear regression is utilized for localizing landmarks. Lee et al. \cite{lee2015face} employs Cascade Gaussian Process Regression Trees (cGPRT) with shape-indexed difference of Gaussian features to achieve face alignment. It has a better generalization ability than cascade regression trees, and shows strong robustness against geometric variations of faces. Most of these methods give an initial shape and refine the shape in an iterative manner, where the final solutions are prone to getting trapped in a local optimum with a poor initialization. In contrast, our method uses a deep neural network to regress from raw face patches to the locations of landmarks.

\subsection{Unconstrained Face Alignment}

Large pose variations and severe occlusions are major challenges in unconstrained environments. Unconstrained face alignment methods are based on 3D models or deal with occlusions explicitly.

Many methods utilize 3D shape models to solve large-pose face alignment. Nair et al. \cite{nair20093} refined the fit of a 3D point distribution model to perform landmark detection. Yu et al. \cite{yu2013pose} used a cascaded deformable shape model to detect landmarks of large-pose faces. Cao et al. \cite{cao2014displaced} employed a displaced dynamic expression regression to estimate the 3D face shape and 2D facial landmarks. The predicted 2D landmarks are used to adjust the model parameters to better fit the current user. Jeni et al. \cite{jeni2015dense} proposed a 3D cascade regression method to implement 3D face alignment, which can maintain the pose invariance of facial landmarks within the range of around $60$ degrees.

There are several occlusion-free face alignment methods. Burgos-Artizzu et al. \cite{burgos2013robust} developed a Robust Cascaded Pose Regression (RCPR) method to detect occlusions explicitly, and uses shape-indexed features to regress the shape increment. Yu et al. \cite{yu2014consensus} utilizes a Bayesian model to merge the estimation results from multiple regressors, in which each regressor is trained to localize facial landmarks with a specific pre-defined facial part being occluded. Wu et al. \cite{wu2015robust} proposed a Robust Facial Landmark Detection (RFLD) method, which uses a robust cascaded regressor to handle complex occlusions and large head poses. To improve the performance of occlusion estimation, landmark visibility probabilities are estimated with an explicit occlusion constraint. Different from these methods, our method is not based on 3D models and does not process occlusions explicitly.

\subsection{Face Alignment via Deep Learning}

Deep learning methods can be divided into two classes: single network based and multiple networks based.

Sun et al. \cite{sun2013deep} estimated the locations of $5$ facial landmarks using Cascaded Convolutional Neural Networks (Cascaded CNN), in which each level computes averaged estimated shape and the shape is refined level by level. Zhou et al. \cite{zhou2013extensive} used multi-level deep networks to detect facial landmarks from coarse to fine. Similarly, Zhang et al. \cite{zhang2014coarse} proposed Coarse-to-Fine Auto-encoder Networks (CFAN). These methods all use multi-stage deep networks to localize landmarks in a coarse-to-fine manner. Instead of using cascaded networks, Honari et al. \cite{honari2016recombinator} proposed Recombinator Networks (RecNet) for learning coarse-to-fine feature aggregation with multi-scale input maps, where each branch extracts features based on current maps and the feature maps of coarser branches.

A few methods employ a single network to solve the face alignment problem. Shao et al. \cite{shao2016learning} proposed a Coarse-to-Fine Training (CFT) method to learn the mapping from input face patches to estimated shapes, which searches the solutions smoothly by adjusting the relative weights between principal landmarks and elaborate landmarks. Zhang et al. \cite{zhang2014facial,zhang2015learning} used the TCDCN with auxiliary facial attribute recognition to predict correlative facial properties like expression and pose, which improves the performance of face alignment. Xiao et al. \cite{xiao2016robust} proposed a Recurrent Attentive-Refinement (RAR) network for face alignment under unconstrained conditions, where shape-indexed deep features and temporal information are taken as inputs and shape predictions are recurrently revised. Compared to these methods, our method uses only one network and is independent of additional facial attributes.

\begin{figure*}[!htb]
\centering\includegraphics[width=17cm]{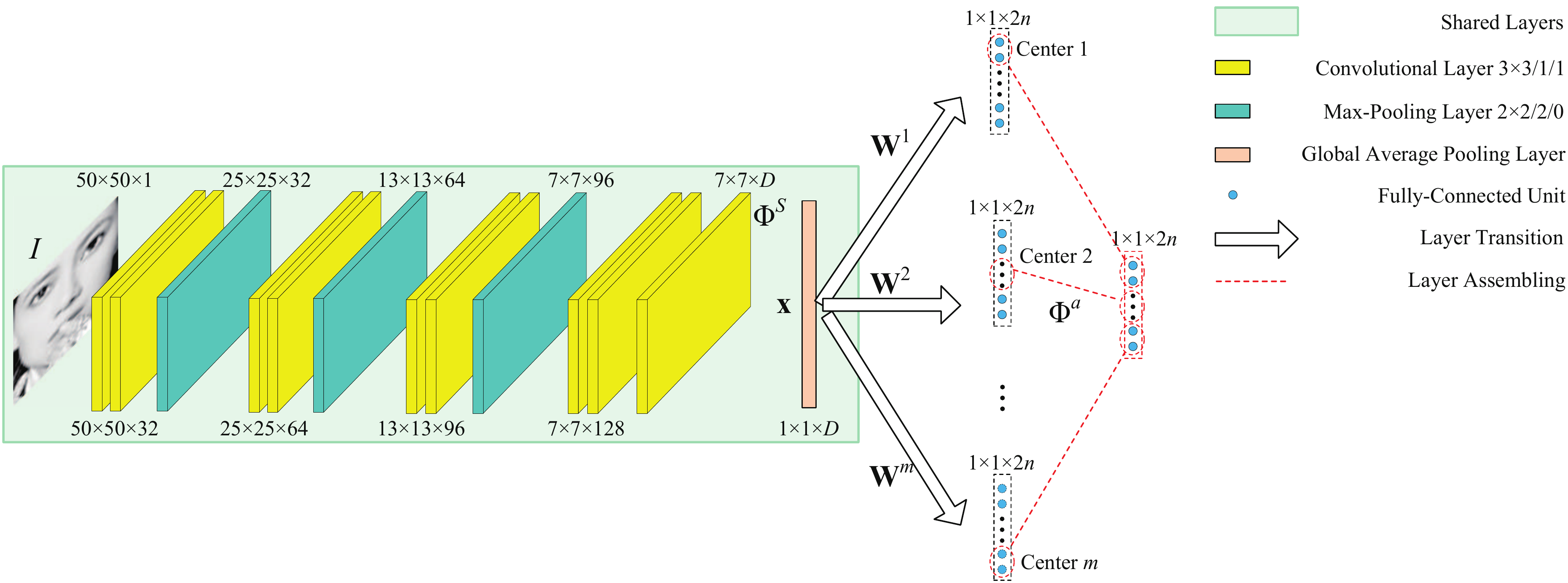}
\caption{Architecture of our network MCL. The expression $k_1 \times k_2 \times k_3$ attached to each layer denotes the height, width, and channel respectively. Every two convolutional layers possess the same expression. The expression $k_4 \times k_5 / k_6 / k_7$ denotes the height, width, stride, and padding of filters respectively. The same type of layers use the identical filters.}
\label{fig:netstruc}
\end{figure*}

\section{Multi-Center Learning for Face Alignment}
\label{sec:mcnet}

\subsection{Network Architecture}
\label{ssec:network}

The architecture of our network MCL is illustrated in Fig. \ref{fig:netstruc}. MCL contains three max-pooling layers, each of which follows a stack of two convolutional layers proposed by VGGNet \cite{simonyan2014very}. In the fourth stack of convolutional layers, we use a convolutional layer with $D$ feature maps above two convolutional layers. We perform Batch Normalization (BN) \cite{ioffe2015batch} and Rectified Linear Unit (ReLU) \cite{nair2010rectified} after each convolution to accelerate the convergence of our network. Most of the existing deep learning methods such as TCDCN \cite{zhang2014facial,zhang2015learning} use the fully-connected layer to extract features, which is apt to overfit and hamper the generalization ability of the network. To sidestep these problems, we operate Global Average Pooling \cite{lin2013network} on the last convolutional layer to extract a high-level feature representation $\mathbf{x}$, which computes the average of each feature map. With this improvement, our MCL acquires a higher representation power with fewer parameters.

Face alignment can be regarded as a nonlinear regression problem, which transforms appearance to shape. A transformation $\Phi^S(\cdot)$ is used for modeling this highly nonlinear function, which extracts the feature $\mathbf{x}$ from the input face image $I$, formulated as
\begin{equation}
\mathbf{x}=\Phi^S(I),
\end{equation}
where $\mathbf{x}=(x_0,x_1,\cdots,x_{D})^T\in \mathbb{R}^{(D+1)\times 1}$, $x_0=1$ corresponds to the bias, and $\Phi^S(\cdot)$ is a composite function of operations including convolution, BN, ReLU, and pooling.

Traditionally, only one shape prediction layer is used, which limits the performance. In contrast, our MCL uses multiple shape prediction layers, each of which emphasizes on the detection of a certain cluster of landmarks. The first several layers are shared by multiple shape prediction layers, which are called shared layers forming the composite function $\Phi^S(\cdot)$.
For the $i$-th shape prediction layer, $i=1,\cdots,m$, a weight matrix $\mathbf{W}^{i}=(\mathbf{w}^{i}_1,\mathbf{w}^{i}_2,\cdots,\mathbf{w}^{i}_{2n})\in \mathbb{R}^{(D+1)\times 2n}$ is used to connect the feature $\mathbf{x}$,  where $m$ and $n$ are the number of shape prediction layers and landmarks, respectively. The reason why we train each shape prediction layer to predict $n$ landmarks instead of one cluster of landmarks is that different facial parts have correlations, as shown in Fig. \ref{fig:unconstrained}.

To decrease the model complexity, we use a model assembling function $\Phi^a(\cdot)$ to integrate multiple shape prediction layers into one shape prediction layer, which is formulated as
\begin{equation}
\label{eq:assemble}
\mathbf{W}^{a}=\Phi^a(\mathbf{W}^{1},\cdots,\mathbf{W}^{m}),
\end{equation}
where $\mathbf{W}^{a}=(\mathbf{w}^{a}_1,\mathbf{w}^{a}_2,\cdots,\mathbf{w}^{a}_{2n})\in \mathbb{R}^{(D+1)\times 2n}$ is the assembled weight matrix. Specifically, $\mathbf{w}^{a}_{2j-1}=\mathbf{w}^{i}_{2j-1}$, $\mathbf{w}^{a}_{2j}=\mathbf{w}^{i}_{2j}$, $j\in P^{i}$, $i=1,\cdots,m$, where $P^{i}$ is the $i$-th cluster of indexes of landmarks. The final prediction $\hat{\mathbf{y}}=(\hat{y}_1,\hat{y}_2,\cdots,\hat{y}_{2n})$ is defined as
\begin{equation}
\label{eq:final_pred}
\hat{\mathbf{y}}={\mathbf{W}^a}^T \mathbf{x},
\end{equation}
where $\hat{y}_{2j-1}$ and $\hat{y}_{2j}$ denote the predicted x-coordinate and y-coordinate of the $j$-th landmark respectively.

Compared to other typical convolutional networks like VGGNet \cite{simonyan2014very}, GoogLe-Net \cite{szegedy2015going}, and ResNet \cite{he2016deep}, our network MCL is substantially smaller and shallower. We believe that such a concise structure is efficient for estimating the locations of facial landmarks. Firstly, face alignment aims to regress coordinates of fewer than $100$ facial landmarks generally, which demands much lower model complexity than visual recognition problems with more than $1,000$ classes. Secondly, a very deep network may fail to work well for landmark detection owing to the reduction of spatial information layer by layer. Other visual localization tasks, like face detection, usually use multiple cascaded shallow networks rather than a single very deep network. Finally, common face alignment benchmarks only contain thousands of training images. A simple network is not easy to overfit given a small amount of raw training data.

\begin{algorithm}[!htb]
\caption{Multi-Center Learning Algorithm.}
\label{alg:learn}
\begin{algorithmic}[1]
\REQUIRE A network MCL, $\Omega^t$, $\Omega^v$, initialized $\Theta$.
\ENSURE $\Theta$.
\STATE Pre-train shared layers and one shape prediction layer until convergence;
\label{alg:learn:basic}
\STATE Fix the parameters of the first six convolutional layers and fine-tune subsequent layers until convergence;
\label{alg:learn:Fix_FT}
\STATE Fine-tune all the layers until convergence;
\label{alg:learn:FT_all}
\FOR{$i=1$ to $m$}
\label{alg:learn:for_beg}
    \STATE Fix $\Theta^{S}$ and fine-tune the $i$-th shape prediction layer until convergence;
    \label{alg:learn:Fix_FT_i}
\ENDFOR
\label{alg:learn:for_end}
\STATE $\Theta=\Theta^{S} \cup \mathbf{W}^{a}$;
\label{alg:learn:combine}
\STATE Return $\Theta$.
\end{algorithmic}
\end{algorithm}

\subsection{Learning Algorithm}
\label{ssec:learning}

The overview of our learning algorithm is shown in Algorithm \ref{alg:learn}. $\Omega^t$ and $\Omega^v$ are the training set and the validation set respectively. $\Theta$ is the set of parameters including weights and biases of our network MCL, which is updated using Mini-Batch Stochastic Gradient Descent (SGD) \cite{krizhevsky2012imagenet} at each iteration. The face alignment loss is defined as
\begin{equation}
\label{eq:loss}
E=\sum_{j=1}^n u_j[(y_{2j-1}-\hat{y}_{2j-1})^2+(y_{2j}-\hat{y}_{2j})^2]/(2d^2),
\end{equation}
where $u_j$ is the weight of the $j$-th landmark, $y_{2j-1}$ and $y_{2j}$ denote the ground-truth x-coordinate and y-coordinate of the $j$-th landmark respectively, and $d$ is the ground truth inter-ocular distance between the eye centers.

Inter-ocular distance normalization provides fair comparisons among faces with different size, and reduces the magnitude of loss to speed up the learning process. During training, a too high learning rate may cause the missing of optimum so far as to the divergence of network, and a too low learning rate may lead to falling into a local optimum. We employ a low initial learning rate to avoid the divergence, and increase the learning rate when the loss is reduced significantly and continue the training procedure.

\subsubsection{Pre-Training and Weighting Fine-Tuning}

In Step \ref{alg:learn:basic}, a \emph{basic model (BM)} with one shape prediction layer is pre-trained to learn a good initial solution. In Eq. \ref{eq:loss}, $u_j=1$ for all $j$. The average alignment error of each landmark of BM on $\Omega^v$ are $\epsilon^b_1, \cdots, \epsilon^b_n$ respectively, which are averaged over all the images. The landmarks with larger errors than remaining landmarks are treated as challenging landmarks.

In Steps \ref{alg:learn:Fix_FT} and \ref{alg:learn:FT_all}, we focus on the detection of challenging landmarks by assigning them larger weights. The weight of the $j$-th landmark is proportional to its alignment error as
\begin{equation}
\label{eq:wm_weight}
u_j=n\epsilon^b_j/\sum_{j=1}^n \epsilon^b_j.
\end{equation}
Instead of fine-tuning all the layers from BM directly, we use two steps to search the solution smoothly. Step \ref{alg:learn:Fix_FT} searches the solution without deviating from BM overly. Step \ref{alg:learn:FT_all} searches the solution within a larger range on the basis of the previous step. This stage is named weighting fine-tuning, which learns a \emph{weighting model (WM)} with higher localization accuracy of challenging landmarks.

\subsubsection{Multi-Center Fine-Tuning and Model Assembling}

\begin{figure}[!htb]
  \centering
  \subfigure[5 landmarks.]{
    \label{fig:face_partition:a} 
    \includegraphics[width=0.9in]{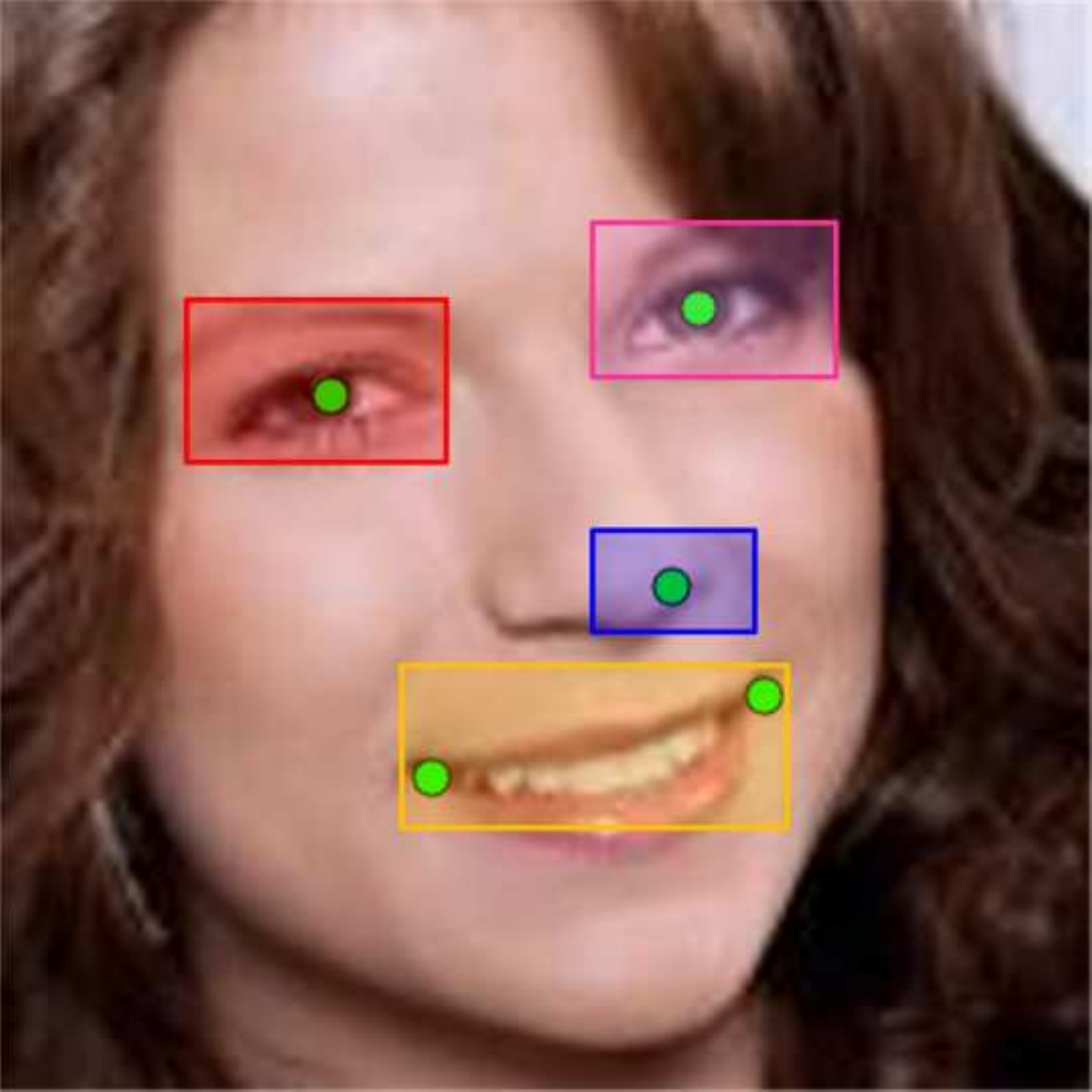}}
  \hspace{0.1in}
  \subfigure[29 landmarks.]{
    \label{fig:face_partition:b} 
    \includegraphics[width=0.9in]{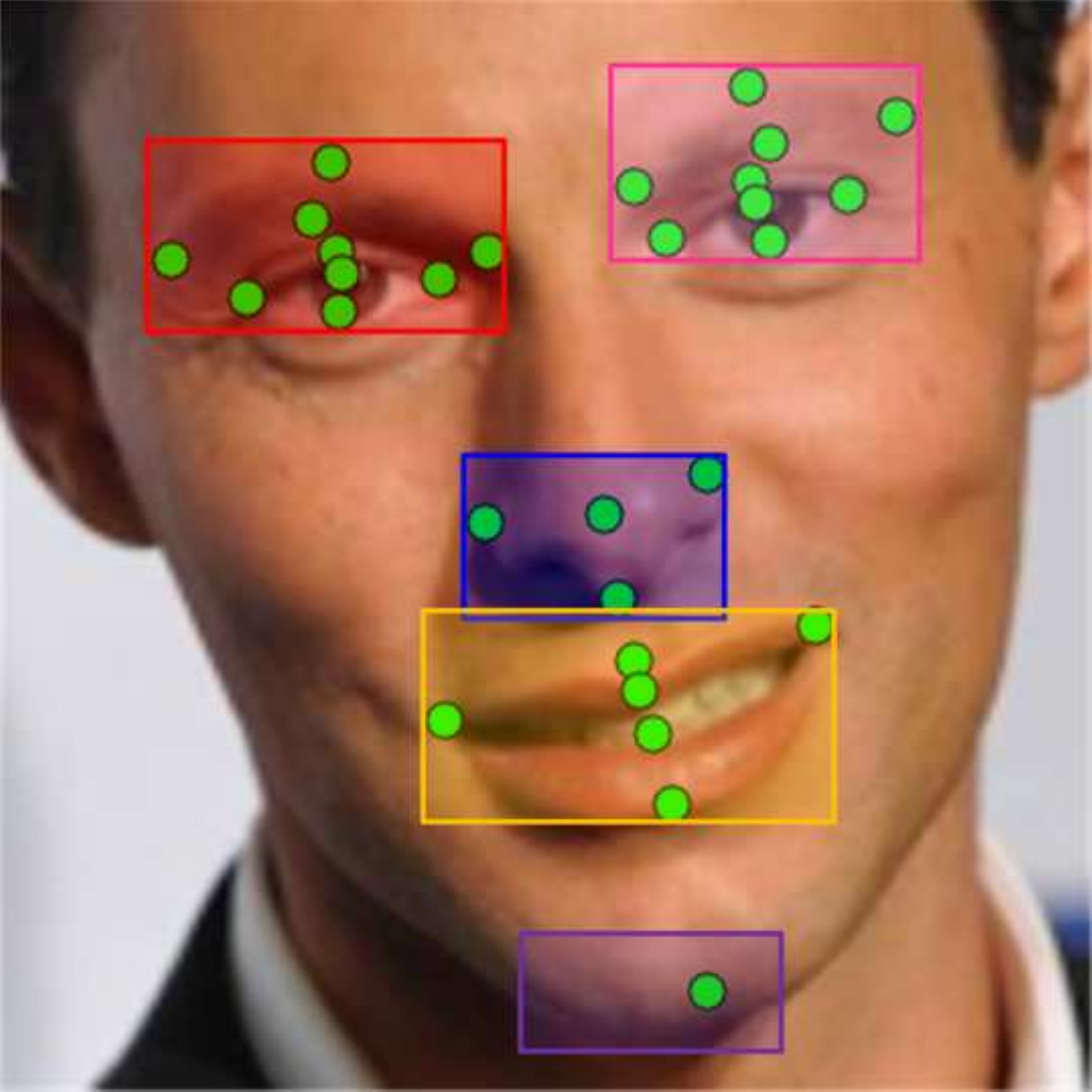}}
  \hspace{0.1in}
  \subfigure[68 landmarks.]{
    \label{fig:face_partition:c} 
    \includegraphics[width=0.9in]{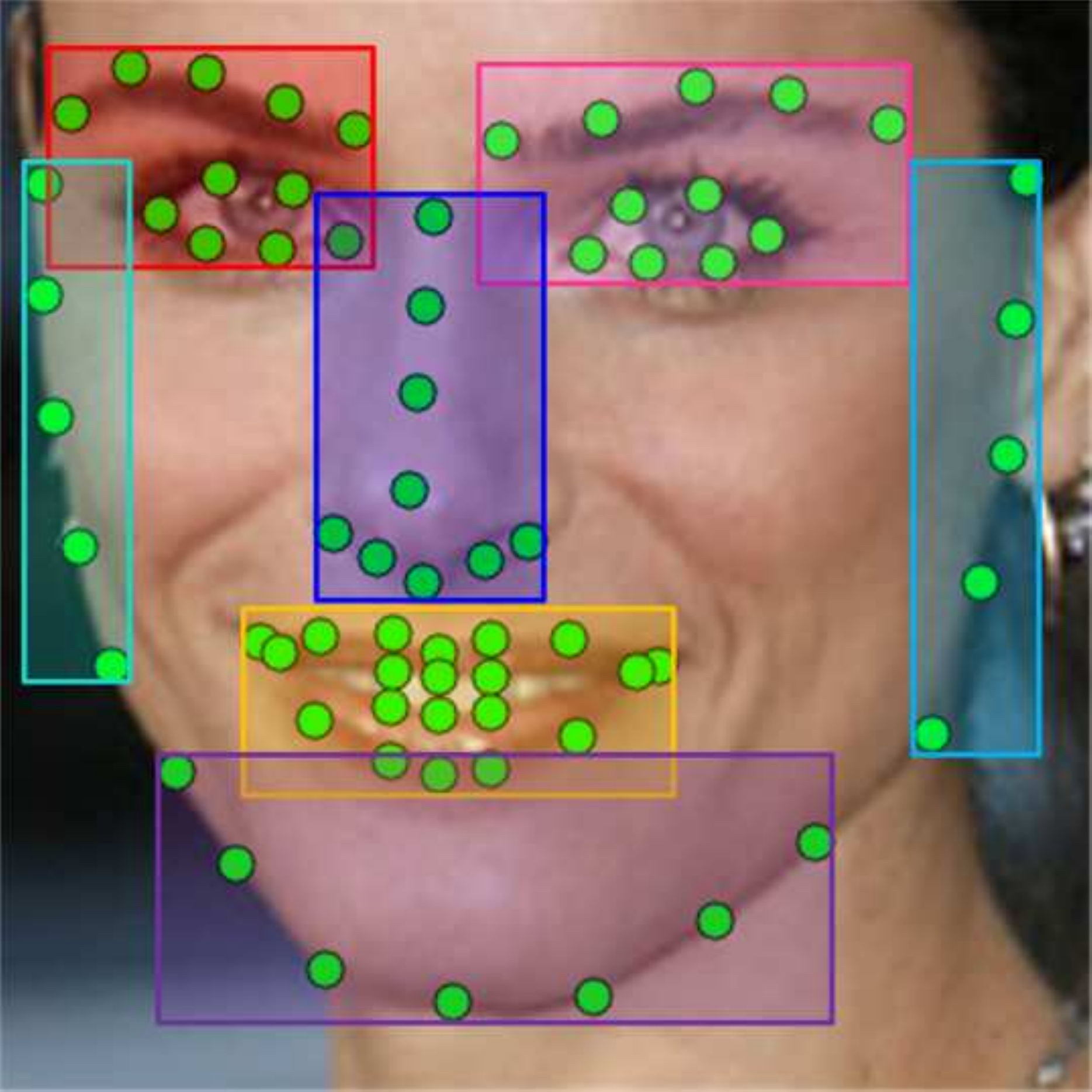}}
  \caption{Partitions of facial landmarks for different labeling patterns.}
  \label{fig:face_partition} 
\end{figure}

The face is partitioned into seven parts according to its semantic structure: left eye, right eye, nose, mouth, left contour, chin, and right contour. As shown in Fig. \ref{fig:face_partition}, different labeling patterns of $5$, $29$, and $68$ facial landmarks are partitioned into $4$, $5$, and $7$ clusters respectively. For the $i$-th shape prediction layer, the $i$-th cluster of landmarks are treated as the optimized center, and the set of indexes of remaining landmarks is denoted as $Q^{i}$.

From Steps \ref{alg:learn:for_beg} to \ref{alg:learn:for_end}, the parameters of shared layers $\Theta^{S}$ are fixed, and each shape prediction layer is initialized with the parameters of the shape prediction layer of WM. When fine-tuning the $i$-th shape prediction layer, the weights of landmarks in $P^{i}$ and $Q^{i}$ are defined as
\begin{equation}
\label{eq:zoom}
u_{P^{i}} = \alpha u_{Q^{i}},
\end{equation}
where $\alpha \gg 1$ is a coefficient to make the $i$-th shape prediction layer emphasize on the detection of the $i$-th cluster of landmarks. The constraint between $u_{P^{i}}$ and $u_{Q^{i}}$ is formulated as
\begin{equation}
\label{eq:sum}
u_{P^{i}} |P^{i}| + u_{Q^{i}} (n-|P^{i}|) = n,
\end{equation}
where $|\cdot|$ refers to the number of elements in a cluster. With Eqs. \ref{eq:zoom} and \ref{eq:sum}, the solved weights are formulated as
\begin{equation}
\begin{split}
&u_{P^{i}} = \alpha n/[(\alpha-1)|P^{i}|+n],\\
&u_{Q^{i}} = n/[(\alpha-1)|P^{i}|+n].
\end{split}
\end{equation}
The average alignment error of each landmark of WM on $\Omega^v$ are $\epsilon^w_1,\cdots, \epsilon^w_n$ respectively. Similar to Eq. \ref{eq:wm_weight}, the weight of the $j$-th landmark is
\begin{eqnarray}
\label{eq:landmark_am}
u_j=
\begin{cases}
u_{P^{i}} |P^{i}| \cdot \epsilon^w_j/\sum_{j\in P^{i}} \epsilon^w_j, &j\in P^{i},\cr u_{Q^{i}} (n-|P^{i}|) \cdot \epsilon^w_j/\sum_{j\in Q^{i}} \epsilon^w_j, &j\in Q^{i}.
\end{cases}
\end{eqnarray}
Although the landmarks in $P^{i}$ are mainly optimized, remaining landmarks are still considered with very small weights rather than zero. This is beneficial for utilizing implicit structural correlations of different facial parts and searching the solutions smoothly. This stage is called multi-center fine-tuning which learns multiple shape prediction layers.

In Step \ref{alg:learn:combine}, multiple shape prediction layers are assembled into one shape prediction layer by Eq. \ref{eq:assemble}. With this model assembling stage, our method learns an \emph{assembling model (AM)}. There is no increase of model complexity in the assembling process, so AM has a low computational cost. It improves the detection precision of each facial landmark by integrating the advantage of each shape prediction layer.

\subsubsection{Analysis of Model Learning}

To investigate the influence from the weights of landmarks on learning procedure, we calculate the derivative of Eq. \ref{eq:loss} with respect to $\hat{y}_k$:
\begin{equation}
\frac{\partial E}{\partial \hat{y}_k}= u_j(\hat{y}_k-y_k)/d^2,
\end{equation}
where $k\in\{2j-1, 2j\}$, $j=1,\cdots,n$. During the learning process, the assembled weight matrix $\mathbf{W}^a$ in Eq. \ref{eq:final_pred} is updated by SGD. Specifically, $\mathbf{w}^{a}_k=\mathbf{w}^{a}_k-\eta \frac{\partial E}{\partial \mathbf{w}^{a}_k} = \mathbf{w}^{a}_k-\eta \frac{\partial E}{\partial \hat{y}_k} \frac{\partial \hat{y}_k}{\partial \mathbf{w}^{a}_k} = \mathbf{w}^{a}_k-\eta \frac{\partial E}{\partial \hat{y}_k} \mathbf{x}$. In summary, $\mathbf{W}^a$ is updated as
\begin{equation}
\mathbf{w}^{a}_k = \mathbf{w}^{a}_k-\eta u_j(\hat{y}_k-y_k) \mathbf{x} /d^2,
\end{equation}
where $\eta$ is the learning rate. If the $j$-th landmark is given a larger weight, its corresponding parameters will be updated with a larger step towards the optimal solution. Therefore, weighting the loss of each landmark ensures that the landmarks with larger weights are mainly optimized. Our method first uses the weighting fine-tuning stage to optimize challenging landmarks, and further uses the multi-center fine-tuning stage to optimize each cluster of landmarks respectively.

\section{Experiments}
\label{sec:experi}

\subsection{Datasets and Settings}

\subsubsection{Datasets}

There are three challenging benchmarks AFLW \cite{kostinger2011annotated}, COFW \cite{burgos2013robust}, and IBUG \cite{sagonas2013300}, which are used for evaluating face alignment with severe occlusion and large variations of pose, expression, and illumination. The provided face bounding boxes are employed to crop face patches during testing.
\begin{itemize}
\item \textbf{AFLW \cite{kostinger2011annotated}} contains $25,993$ faces under real-world conditions gathered from Flickr. Compared with other datasets like MUCT \cite{milborrow2010muct} and LFPW \cite{belhumeur2013localizing}, AFLW exhibits larger pose variations and extreme partial occlusions. Following the settings of \cite{zhang2014facial,honari2016recombinator}, $2,995$ images are used for testing, and $10,000$ images annotated with $5$ landmarks are used for training, which includes $4,151$ LFW \cite{huang2007labeled} images and $5,849$ web images.

\item \textbf{COFW \cite{burgos2013robust}} is an occluded face dataset in the wild, in which the faces are designed with severe occlusions using accessories and interactions with objects. It contains $1,007$ images annotated with $29$ landmarks. The training set includes $845$ LFPW faces and $500$ COFW faces, and the testing set includes remaining $507$ COFW faces.

\item \textbf{IBUG \cite{sagonas2013300}} contains $135$ testing images which present large variations in pose, expression, illumination, and occlusion. The training set consists of AFW \cite{zhu2012face}, the training set of LFPW, and the training set of Helen \cite{le2012interactive}, which are from 300-W \cite{sagonas2013300} with $3,148$ images labeled with $68$ landmarks.
\end{itemize}

\subsubsection{Implementation Details}
\label{sssec:implementation}

We enhance the diversity of raw training data on account of their limited variation patterns, using five steps: rotation, uniform scaling, translation, horizontal flip, and JPEG compression. In particular, for each training face, we firstly perform multiple rotations, and attain a tight face bounding box covering the ground truth locations of landmarks of each rotated result respectively. Uniform scaling and translation with different extents on face bounding boxes are further conducted, in which each newly generated face bounding box is used to crop the face. Finally training samples are augmented through horizontal flip and JPEG compression. It is beneficial for avoiding overfitting and improving the robustness of learned models by covering various patterns.

We train our MCL using an open source deep learning framework Caffe \cite{jia2014caffe}. The input face patch is a $50\times50$ grayscale image, and each pixel value is normalized to $[-1,1)$ by subtracting $128$ and multiplying $0.0078125$. A more complex model is needed for a labeling pattern with more facial landmarks, so $D$ is set to be $512 / 512 / 1,024$ for $5 / 29 / 68$ facial landmarks. The type of solver is SGD with a mini-batch size of $64$, a momentum of $0.9$, and a weight decay of $0.0005$. The maximum learning iterations of pre-training and each fine-tuning step are $18\times10^4$ and $6\times10^4$ respectively, and the initial learning rates of pre-training and each fine-tuning step are $0.02$ and $0.001$ respectively. Note that the initial learning rate of fine-tuning should be low to preserve some representational structures learned in the pre-training stage and avoid missing good intermediate solutions. The learning rate is multiplied by a factor of $0.3$ at every $3\times10^4$ iterations, and the remaining parameter $\alpha$ is set to be $125$.

\subsubsection{Evaluation Metric}
\label{sssec:metric}

Similar to previous methods \cite{cao2012face,sun2013deep,zhang2015learning}, we report the inter-ocular distance normalized mean error, and treat the mean error larger than $10\%$ as a failure. To conduct a more comprehensive comparison, the cumulative errors distribution (CED) curves are plotted. To measure the time efficiency, the average running speed (Frame per Second, FPS) on a single core i5-6200U 2.3GHz CPU is also reported. A single image is fed into the model at a time when computing the speed. In other words, we evaluate methods on four popular metrics: mean error (\%), failure rate (\%), CED curves, and average running speed. In the next sections, \% in all the results are omitted for simplicity.

\subsection{Comparison with State-of-the-Art Methods}
\label{ssec:compare}

We compare our work MCL\footnotetext[2]{The result is acquired by running the code at https://github.com/seetaface/SeetaFaceEngine/tree/master/FaceAlignment.} against state-of-the-art methods including ESR \cite{cao2012face}, SDM \cite{xiong2013supervised}, Cascaded CNN \cite{sun2013deep}, RCPR \cite{burgos2013robust}, CFAN \cite{zhang2014coarse}, LBF \cite{ren2014face}, cGPRT \cite{lee2015face}, CFSS \cite{zhu2015face}, TCDCN \cite{zhang2014facial,zhang2015learning}, ALR \cite{shao2016face}, CFT \cite{shao2016learning}, RFLD \cite{wu2015robust}, RecNet \cite{honari2016recombinator}, RAR \cite{xiao2016robust}, and FLD+PDE \cite{wu2017simultaneous}. All the methods are evaluated on testing images using the face bounding boxes provided by benchmarks. In addition to given training images, TCDCN uses outside training data labeled with facial attributes. RAR augments training images with occlusions incurred by outside natural objects like sunglasses, phones, and hands. FLD+PDE performs facial landmark detection, pose and deformation estimation simultaneously, in which the training data of pose and deformation estimation are used. Other methods including our MCL only utilize given training images from the benchmarks.

\begin{table}[!htb]
\centering\caption{Comparison of results of mean error with state-of-the-art methods. Several methods did not share their results or code on some benchmarks, so we use results marked with ``*'' from \cite{zhang2014facial,zhang2015learning}.}
\label{tab:comp_other_tab}
\begin{tabular}{|*{4}{c|}}
\hline
\multirow{2}*{Method} &AFLW &COFW &IBUG \\
&5 landmarks&29 landmarks&68 landmarks\\
\hline
ESR \cite{cao2012face} &12.4* &11.2* &17.00* \\
SDM \cite{xiong2013supervised} &8.5* &11.14* &15.40* \\
Cascaded CNN \cite{sun2013deep} &8.72 &- &- \\
RCPR \cite{burgos2013robust} &11.6* &8.5 &17.26* \\
CFAN \cite{zhang2014coarse} &7.83\footnotemark[2] &- &16.78* \\
LBF \cite{ren2014face} &- &- &11.98 \\
cGPRT \cite{lee2015face} &- &- &11.03 \\
CFSS \cite{zhu2015face} &- &- &9.98 \\
TCDCN \cite{zhang2014facial,zhang2015learning} &8.0 &8.05 &8.60 \\
ALR \cite{shao2016face} &7.42 &- &- \\
CFT \cite{shao2016learning} &- &6.33 &10.06 \\
RFLD \cite{wu2015robust} &- &\textbf{5.93} &- \\
RecNet \cite{honari2016recombinator} &5.60 &- &8.44 \\
RAR \cite{xiao2016robust} &7.23 &6.03 &\textbf{8.35} \\
FLD+PDE \cite{wu2017simultaneous} &- &6.40 &- \\
\textbf{MCL} &\textbf{5.38} &6.00 &8.51 \\
\hline
\end{tabular}
\end{table}

\begin{figure*}[!htb]
\centering
\subfigure[Results of Cascaded CNN, ALR, and MCL on AFLW.]{
\label{fig:compare results:a}
\begin{minipage}[b]{0.38\textwidth}
\includegraphics[scale=0.11]{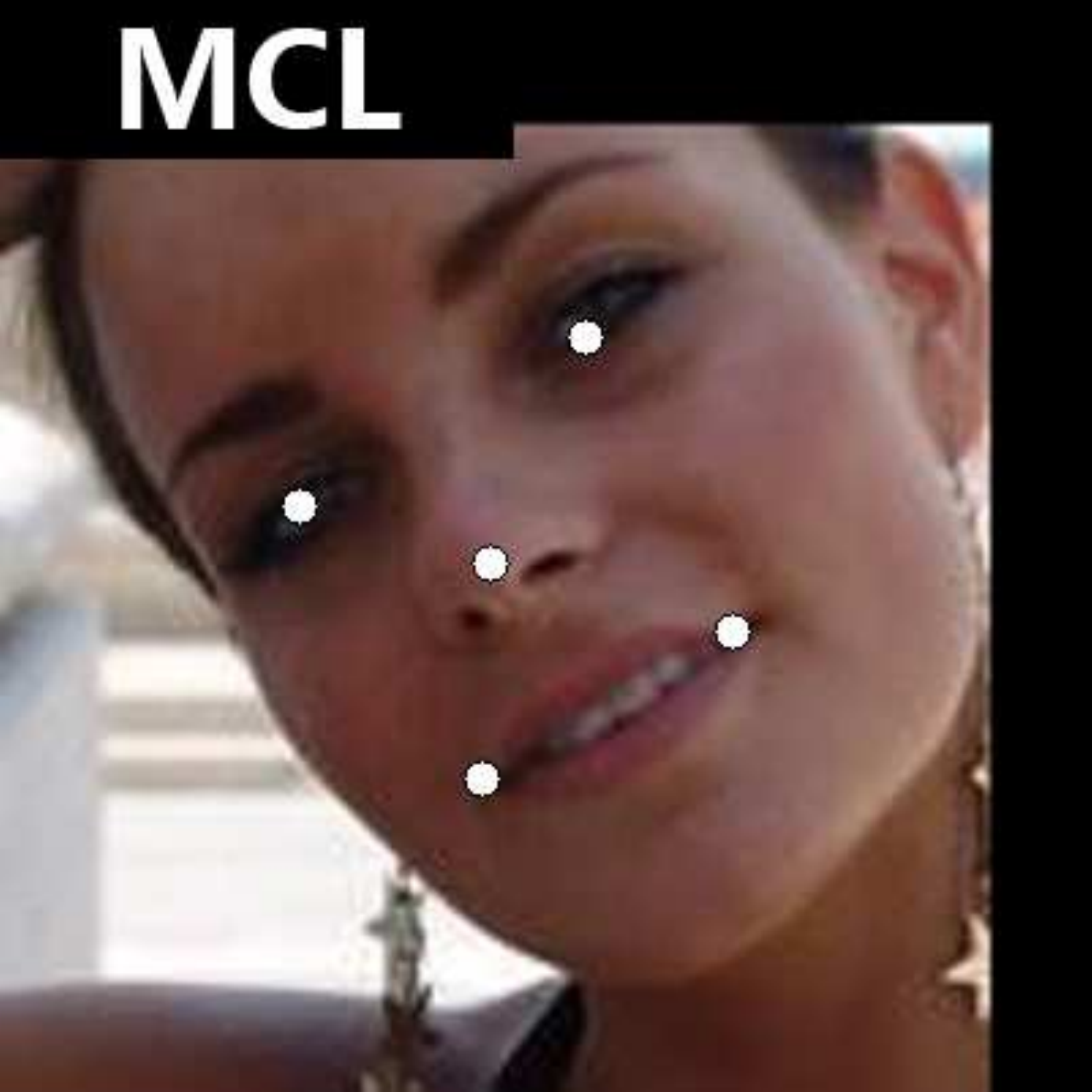}
\includegraphics[scale=0.22]{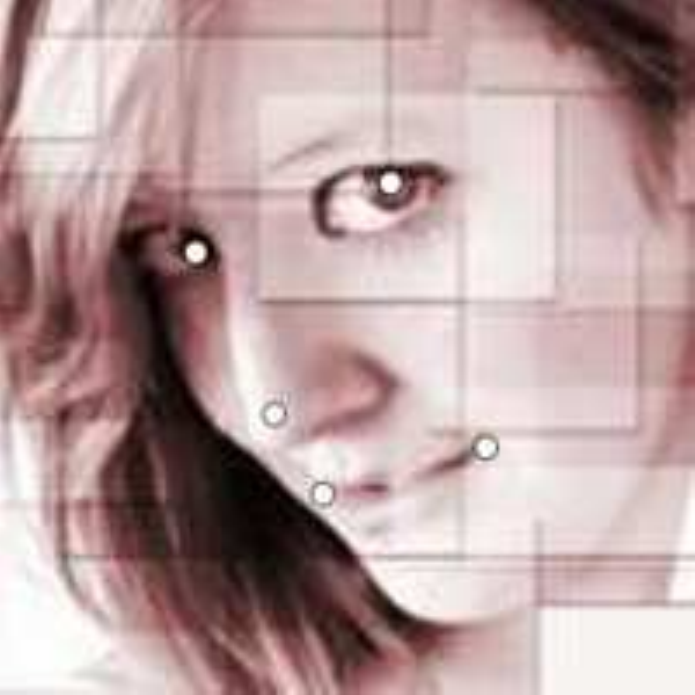}
\includegraphics[scale=0.22]{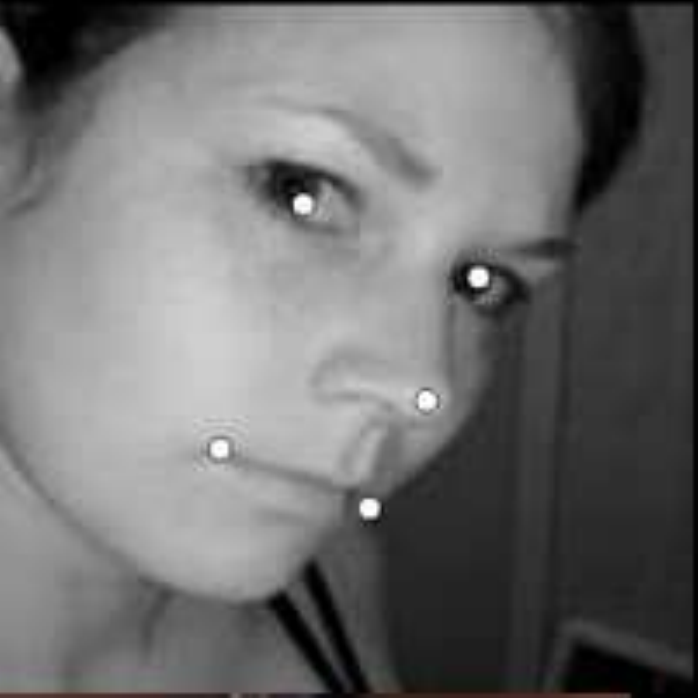}
\includegraphics[scale=0.22]{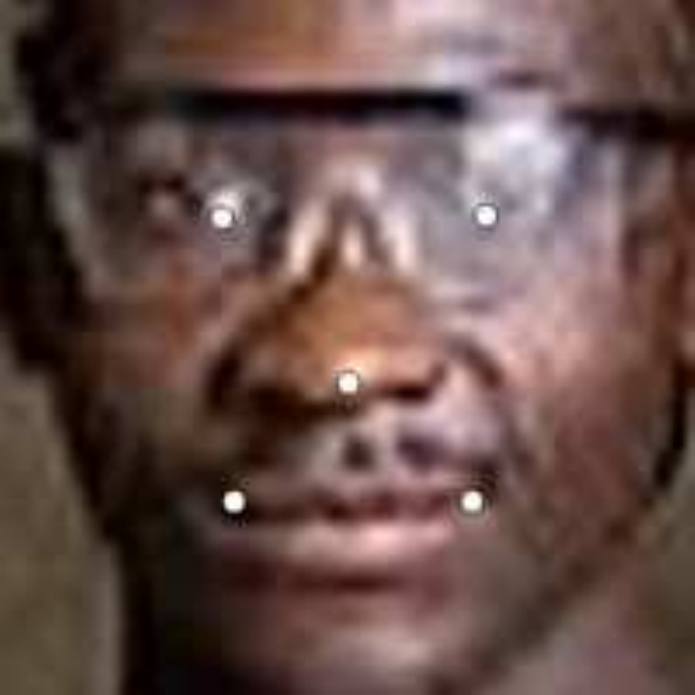}

\includegraphics[scale=0.22]{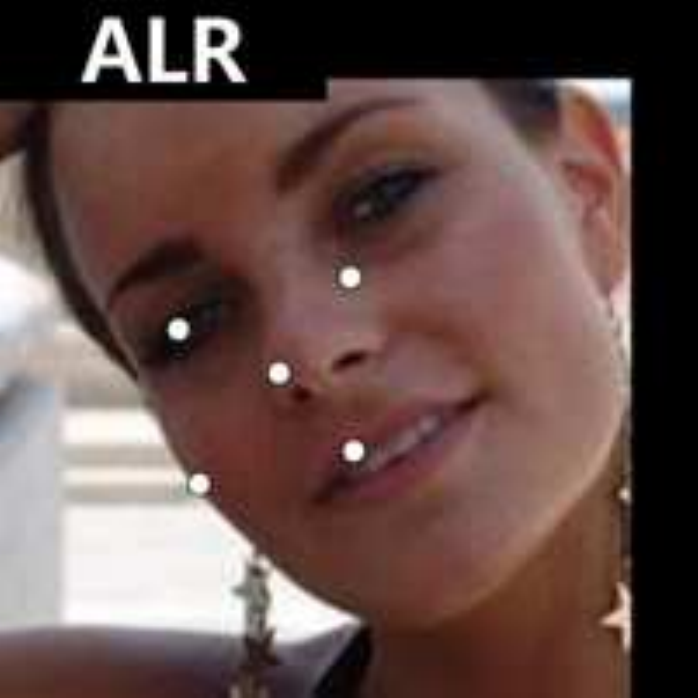}
\includegraphics[scale=0.22]{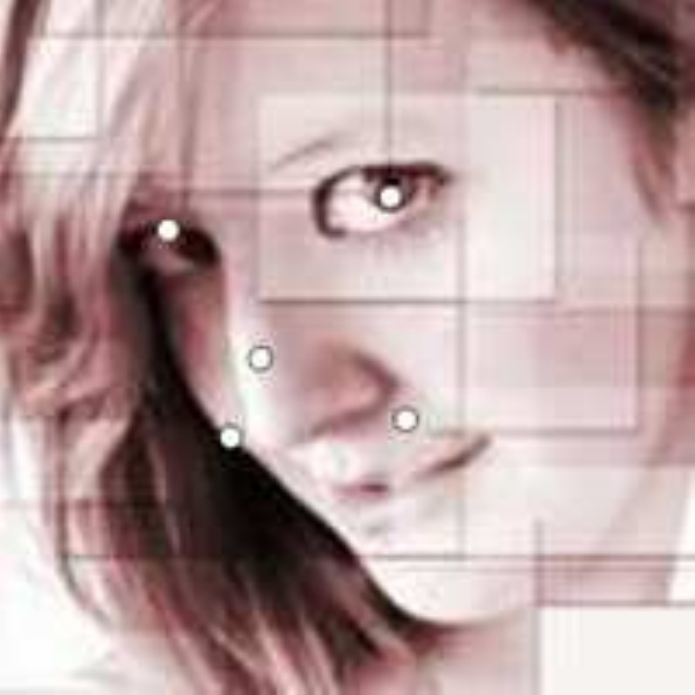}
\includegraphics[scale=0.22]{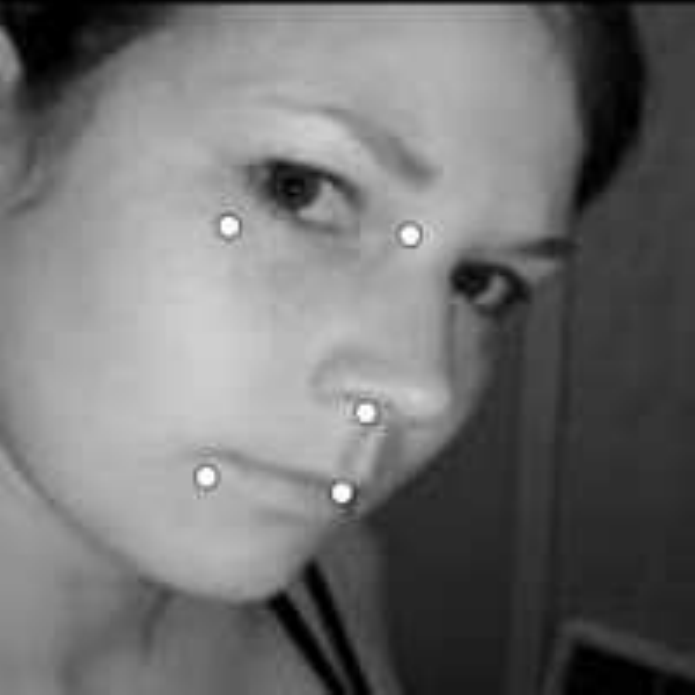}
\includegraphics[scale=0.22]{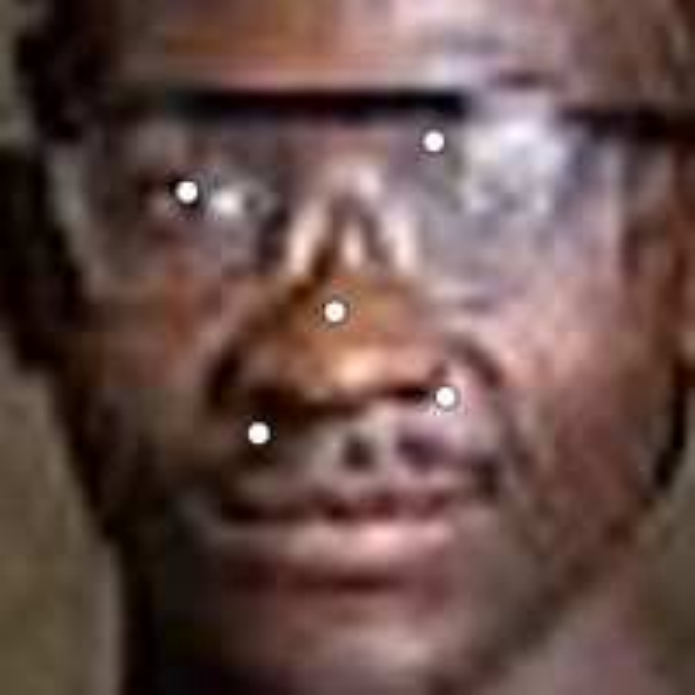}

\includegraphics[scale=0.22]{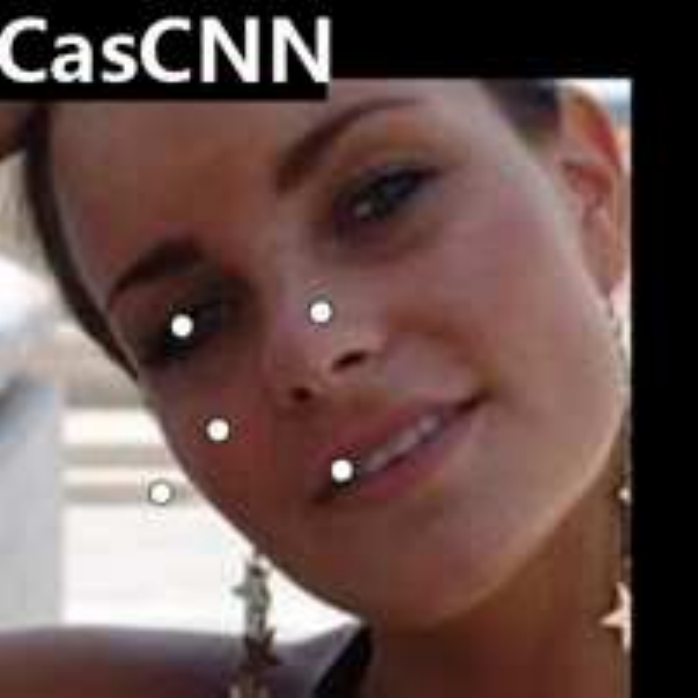}
\includegraphics[scale=0.22]{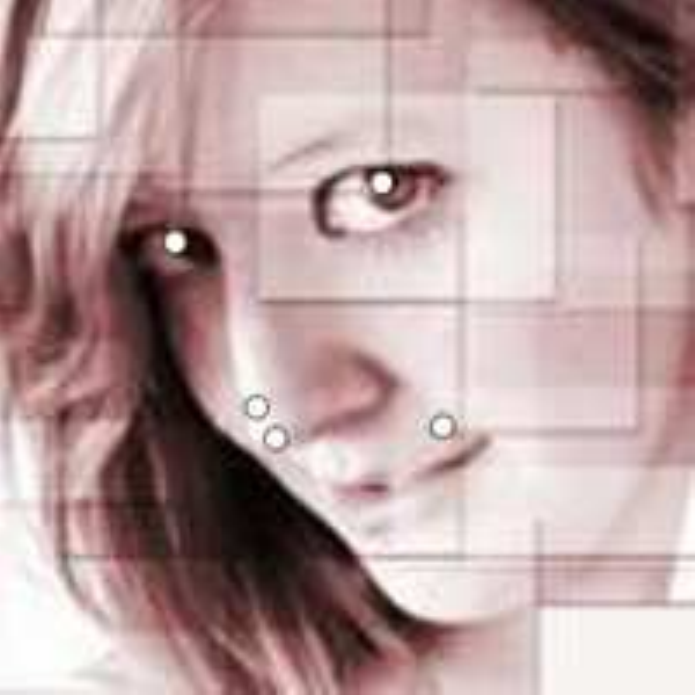}
\includegraphics[scale=0.22]{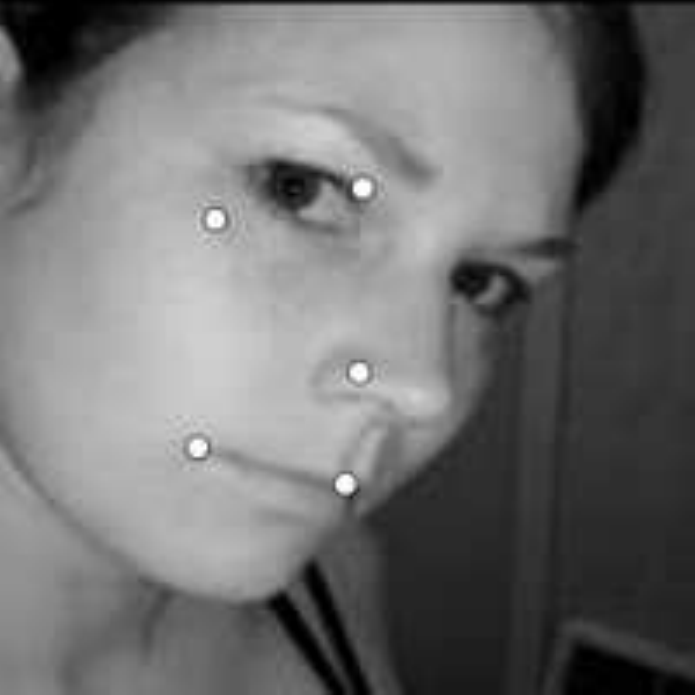}
\includegraphics[scale=0.22]{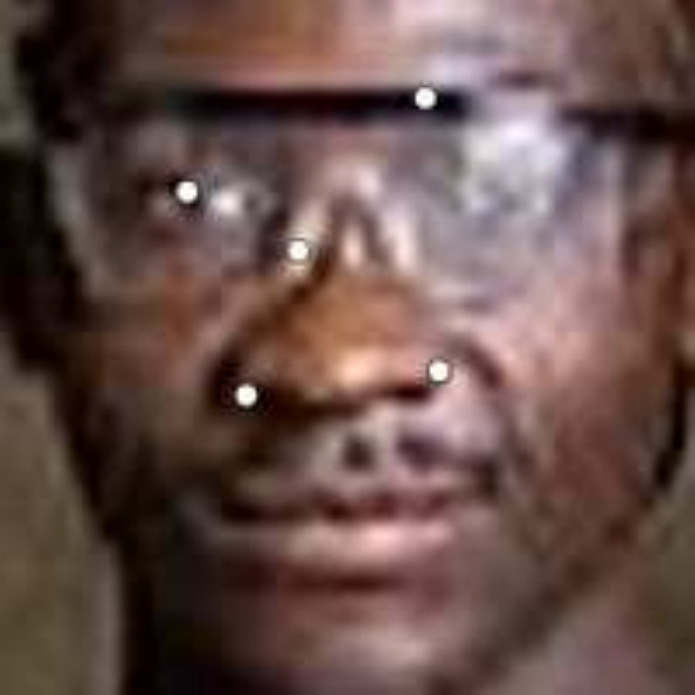}
\end{minipage}
}
\hspace{0in}
\subfigure[Results of RCPR, CFT, and MCL on COFW.]{
\label{fig:compare results:b}
\begin{minipage}[b]{0.38\textwidth}
\includegraphics[scale=0.11]{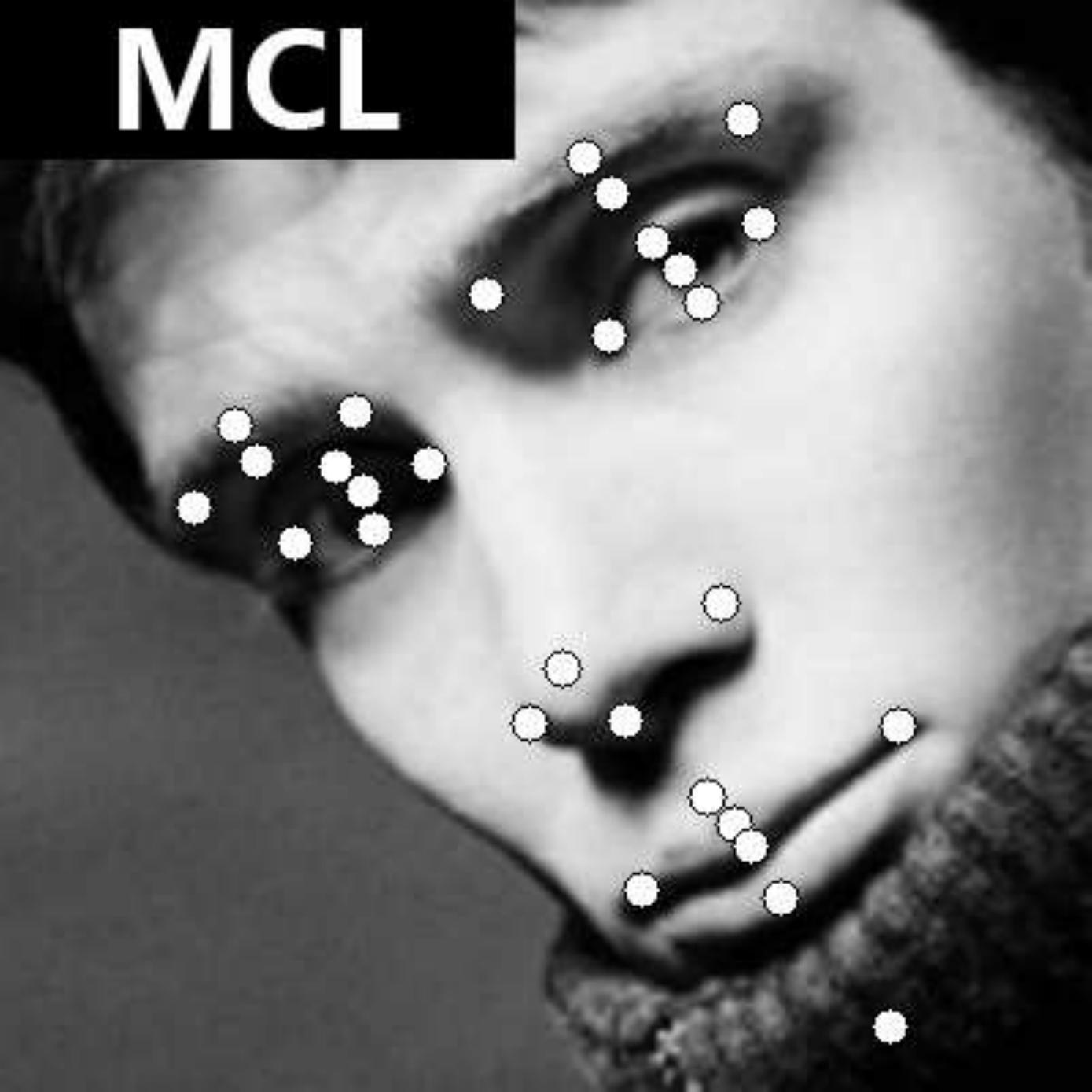}
\includegraphics[scale=0.22]{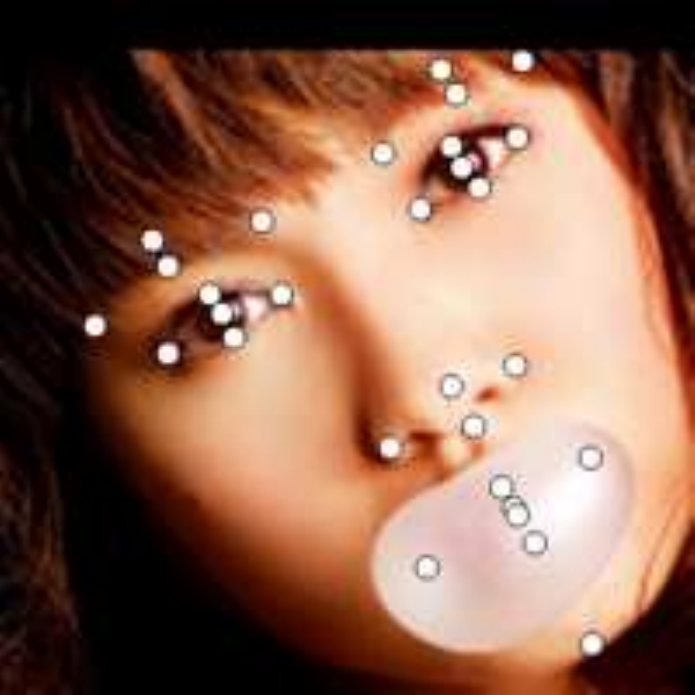}
\includegraphics[scale=0.22]{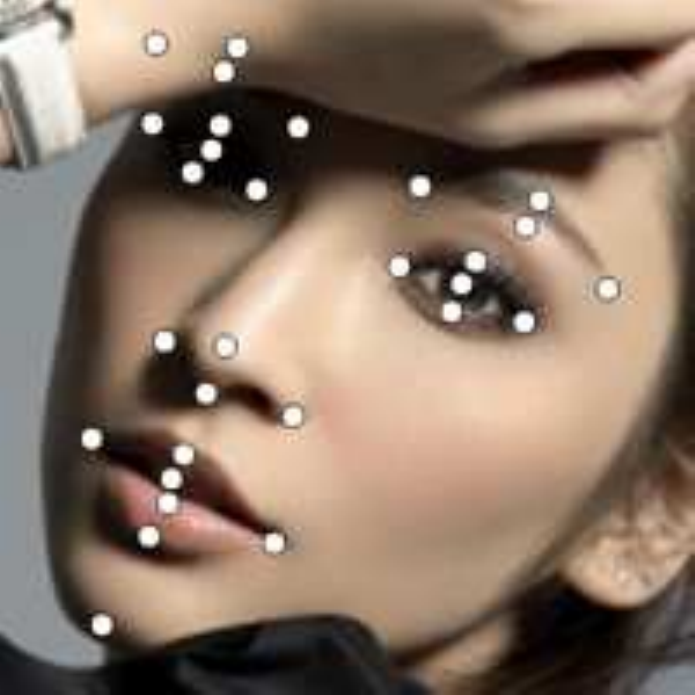}
\includegraphics[scale=0.22]{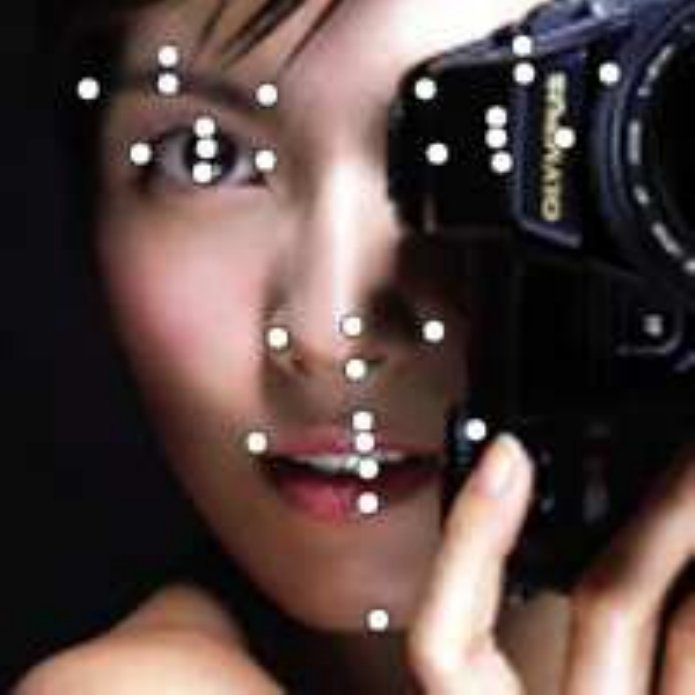}

\includegraphics[scale=0.22]{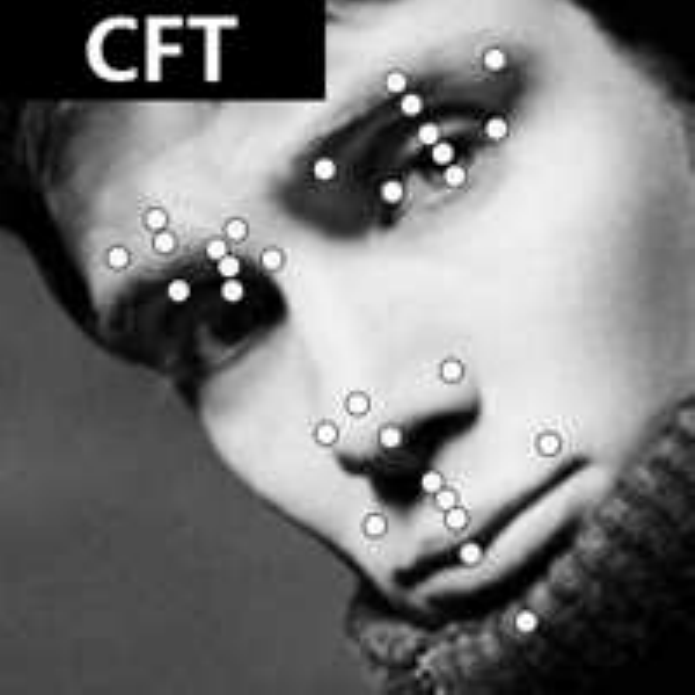}
\includegraphics[scale=0.22]{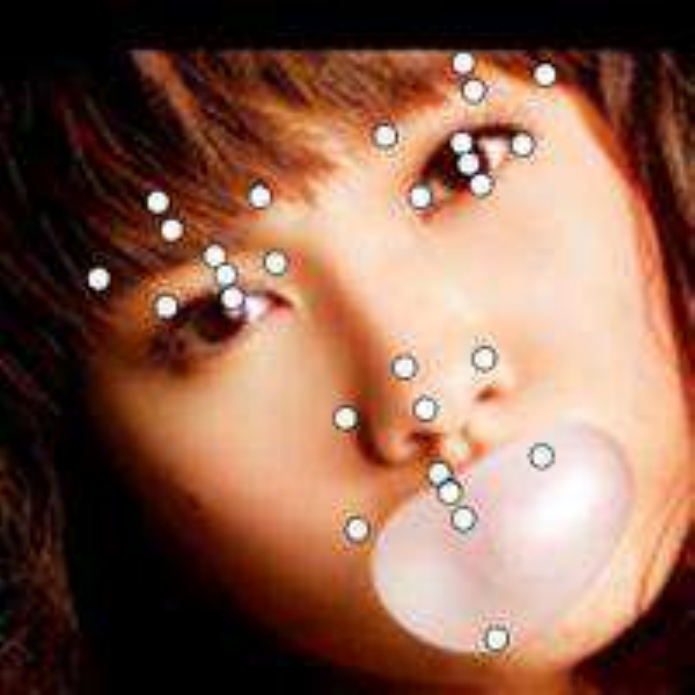}
\includegraphics[scale=0.22]{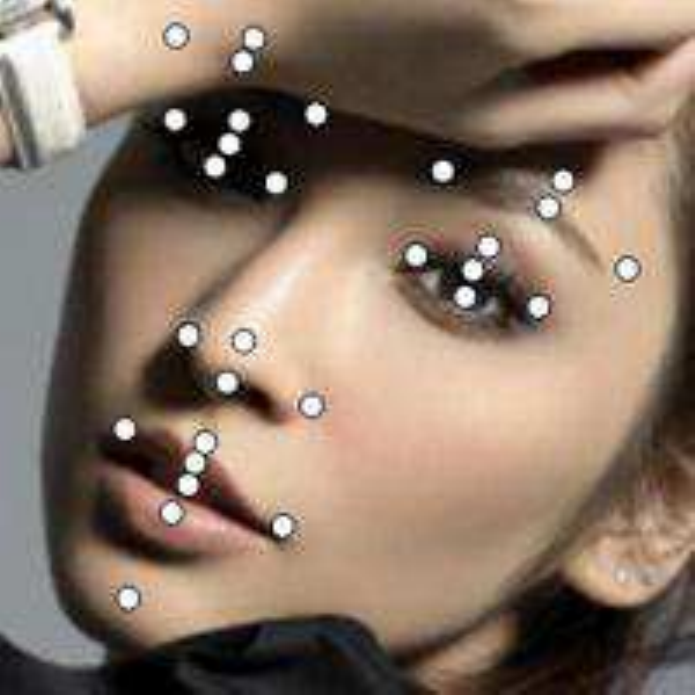}
\includegraphics[scale=0.22]{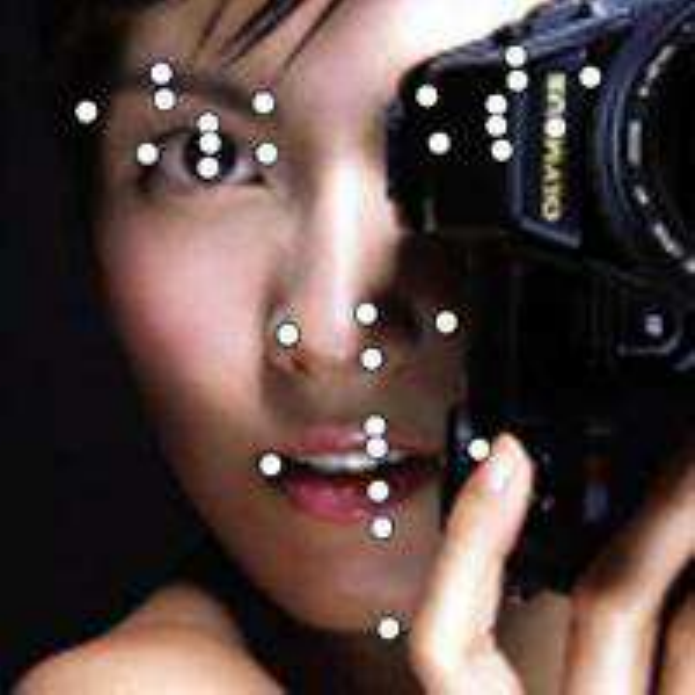}

\includegraphics[scale=0.22]{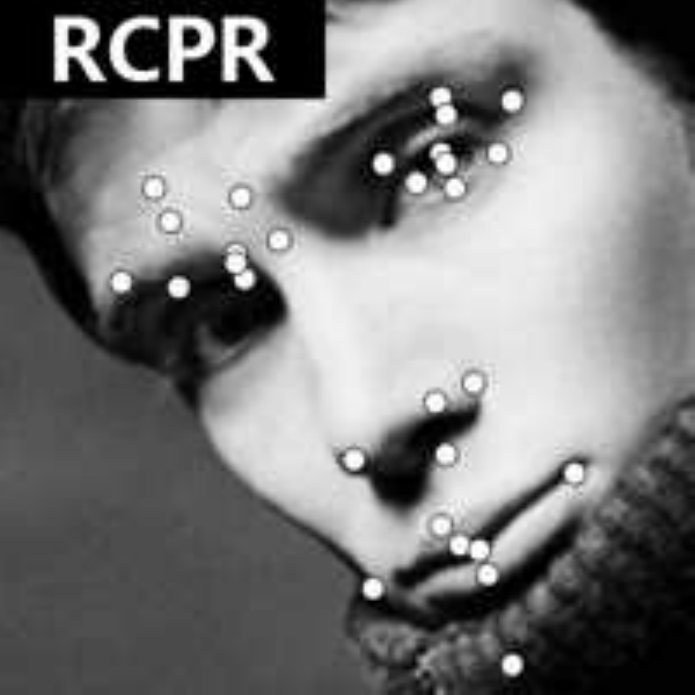}
\includegraphics[scale=0.22]{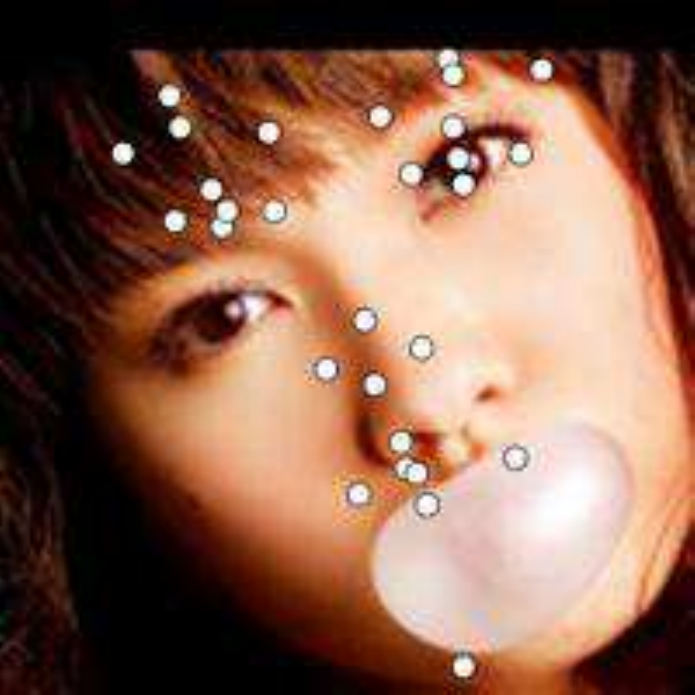}
\includegraphics[scale=0.22]{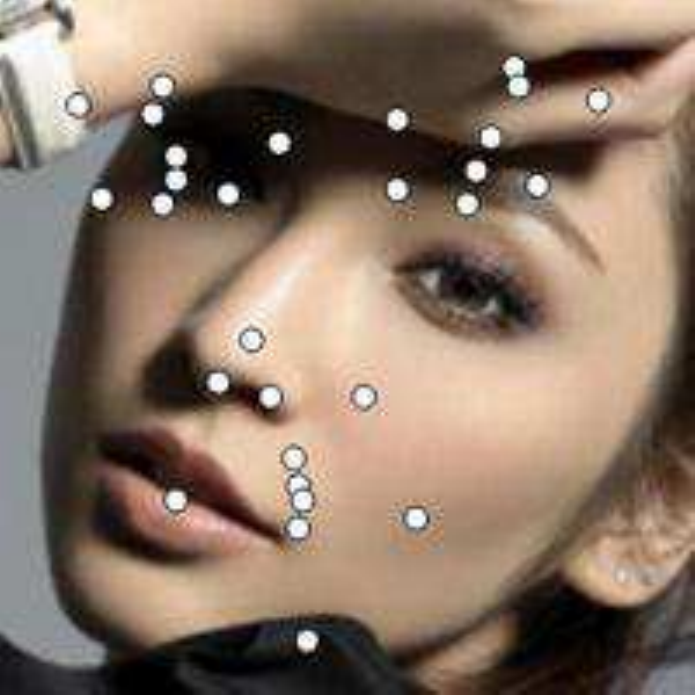}
\includegraphics[scale=0.22]{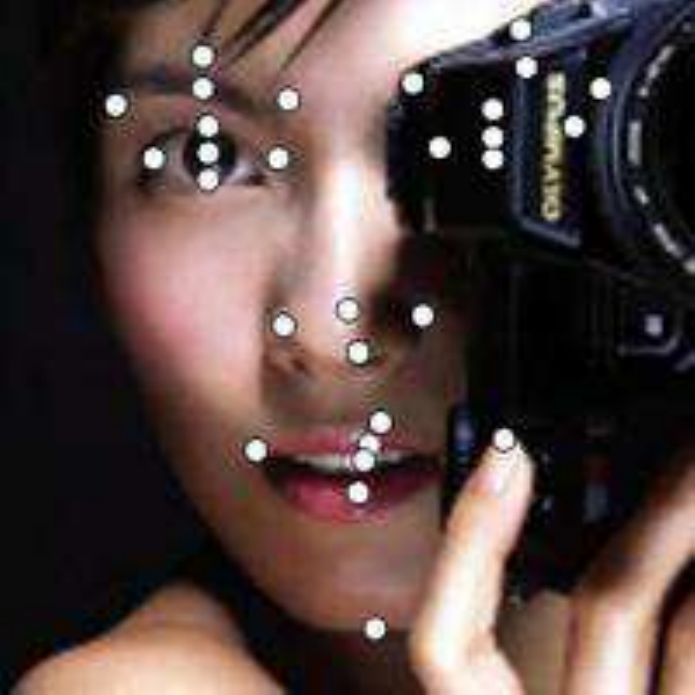}
\end{minipage}
}
\caption{Face alignment results of state-of-the-art methods and our method MCL on challenging faces.}
\label{fig:compare results}
\end{figure*}

Table \ref{tab:comp_other_tab} reports the results of our method and previous works on three benchmarks. Our method MCL outperforms most of the state-of-the-art methods, especially on AFLW dataset where a relative error reduction of $3.93\%$ is achieved compared to RecNet. Cascaded CNN estimates the location of each landmark separately in the second and third level, and every two networks are used to detect one landmark. It is difficult to be extended to dense landmarks owing to the explosion of the number of networks. TCDCN relies on outside training data for auxiliary facial attribute recognition, which limits the universality. It can be seen that MCL outperforms Cascaded CNN and TCDCN on all the benchmarks. Moreover, MCL is robust to occlusions with the performance on par with RFLD, benefiting from utilizing semantical correlations among different landmarks. RecNet and RAR show significant results, but their models are very complex with high computational costs.

\begin{figure*}[!htb]
\centering
\includegraphics[scale=0.11]{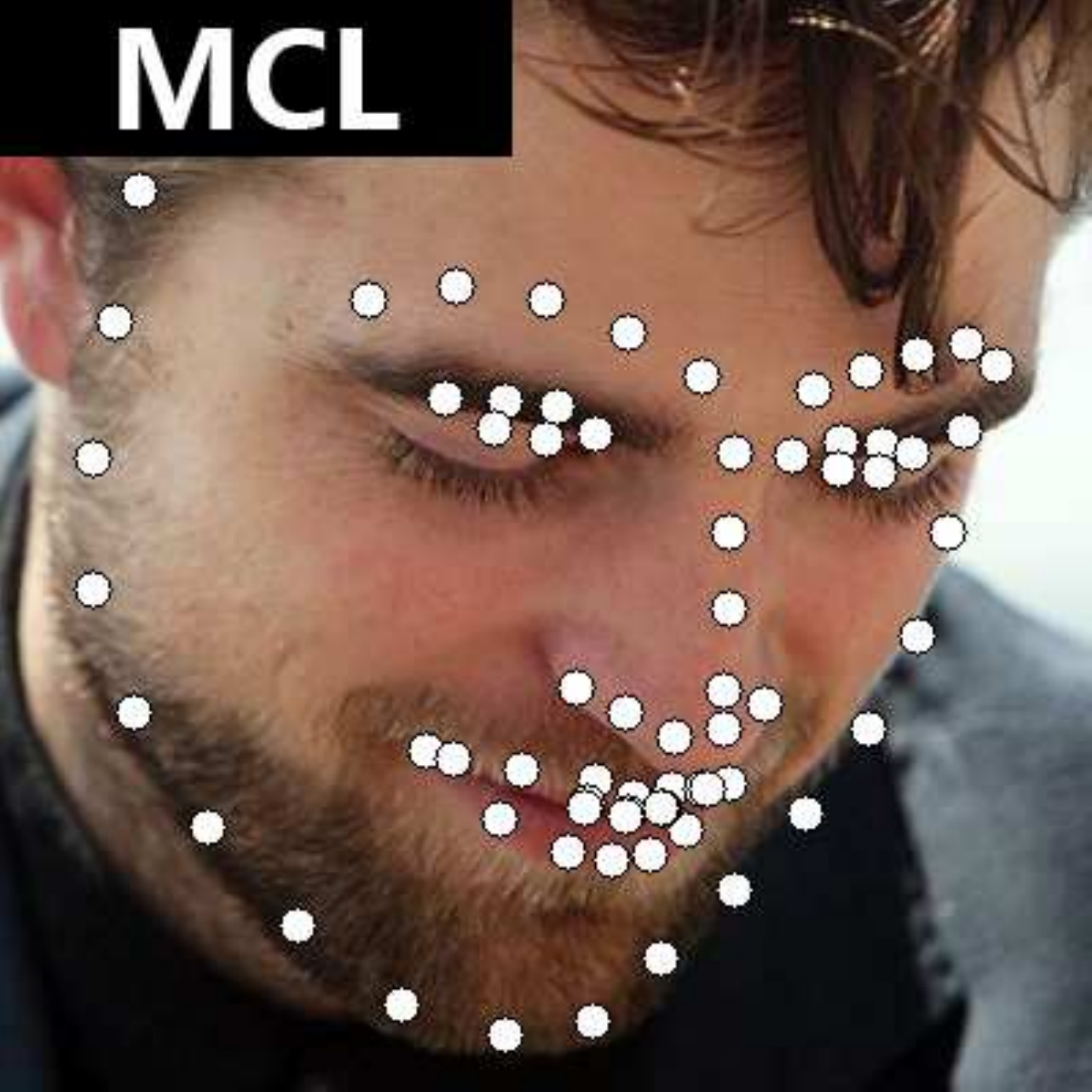}
\includegraphics[scale=0.22]{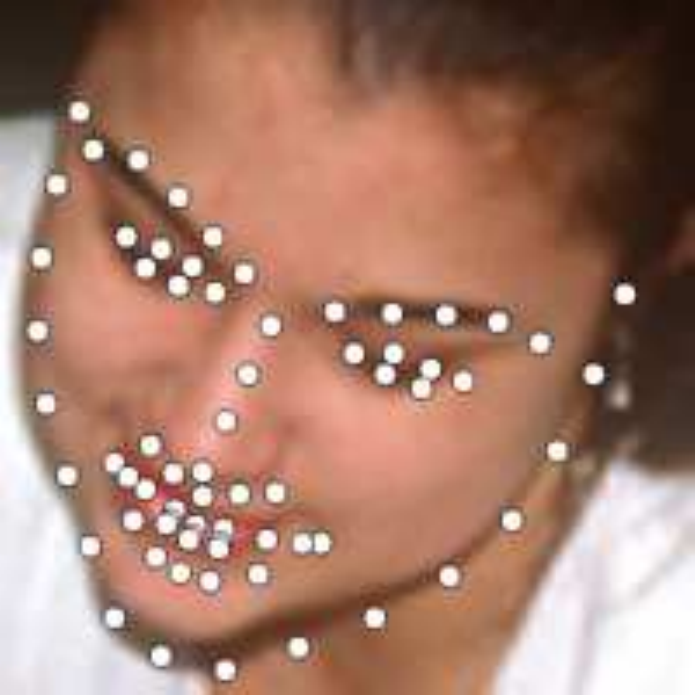}
\includegraphics[scale=0.22]{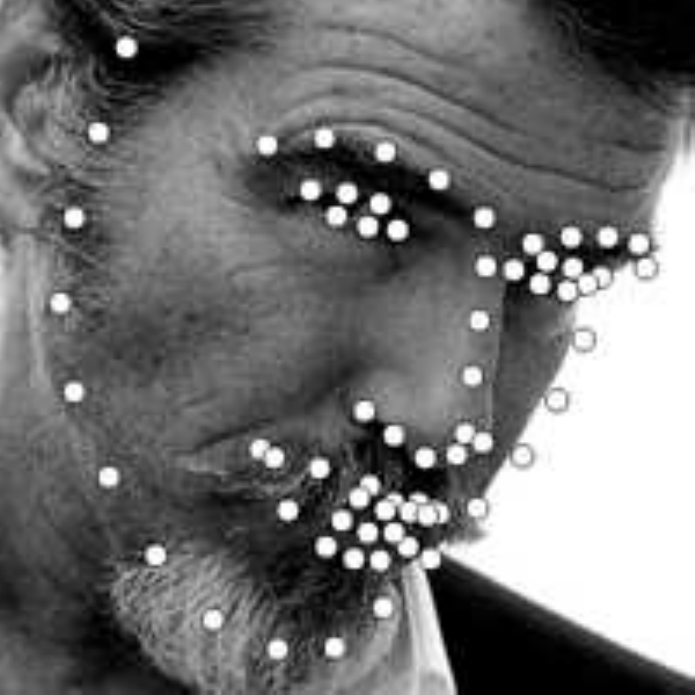}
\includegraphics[scale=0.22]{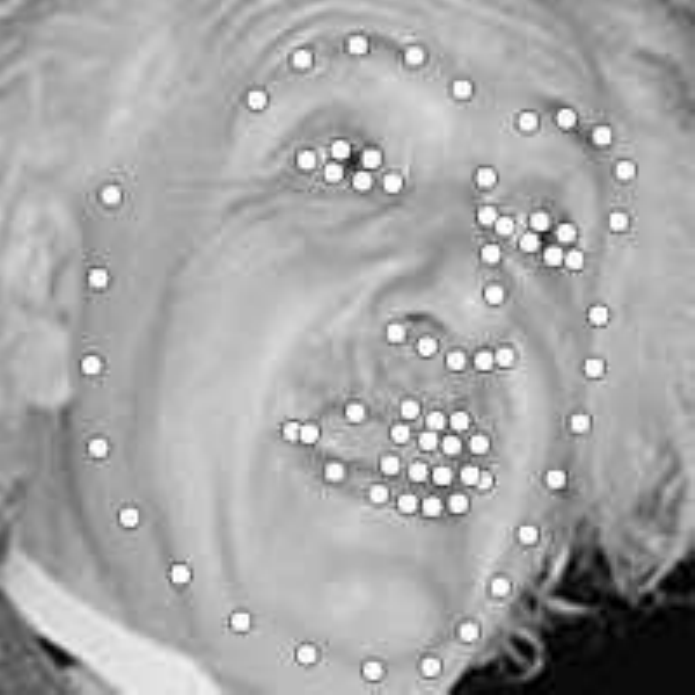}
\includegraphics[scale=0.22]{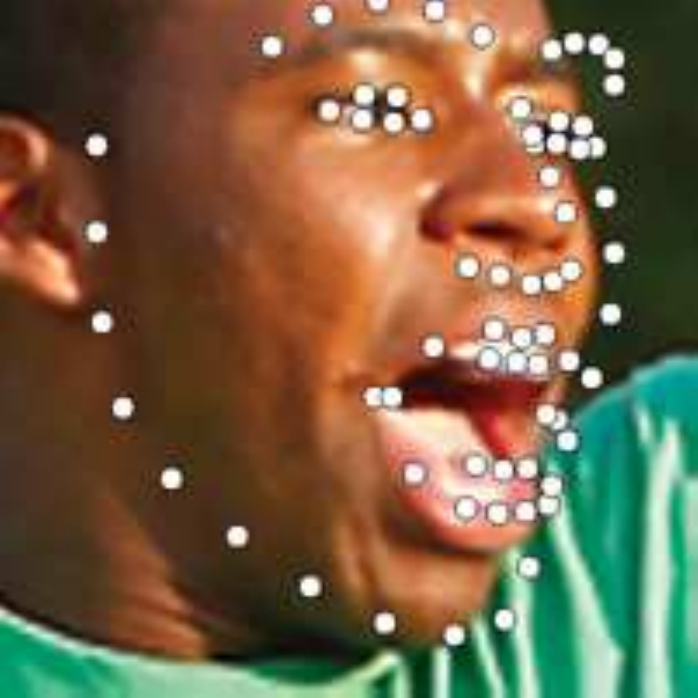}
\includegraphics[scale=0.22]{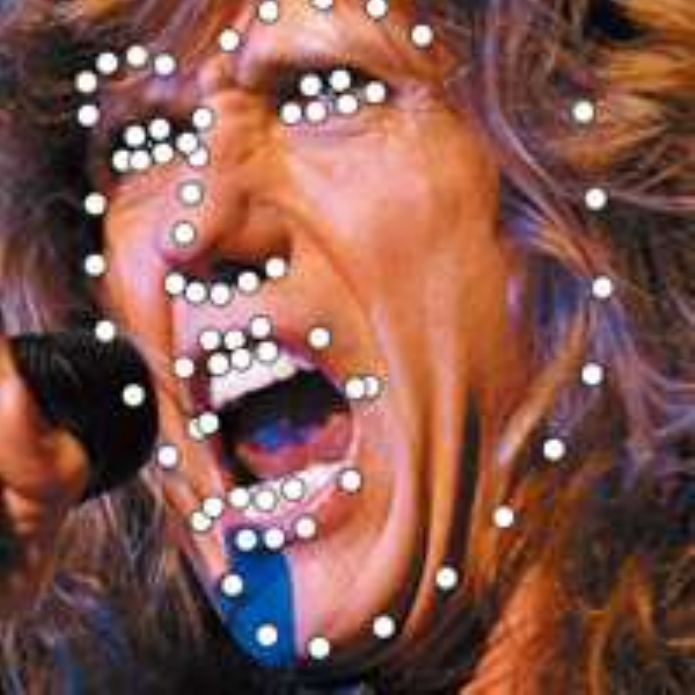}
\includegraphics[scale=0.22]{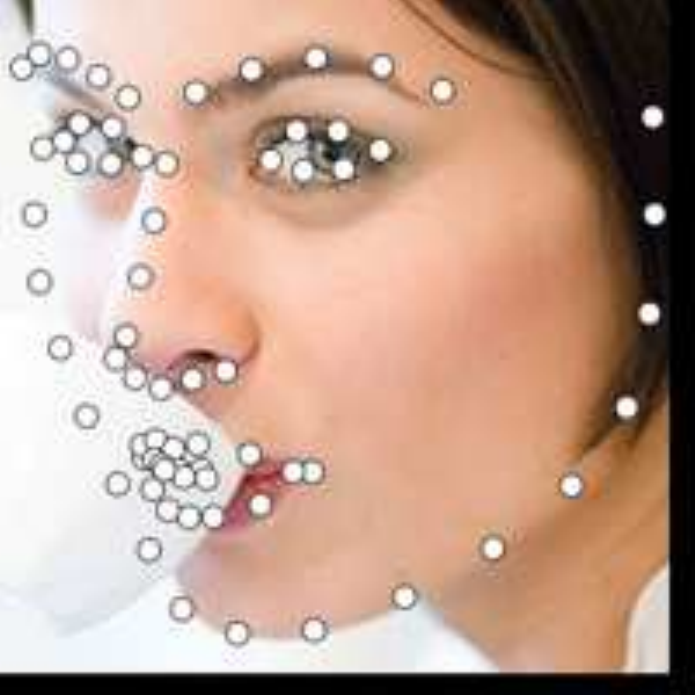}
\includegraphics[scale=0.22]{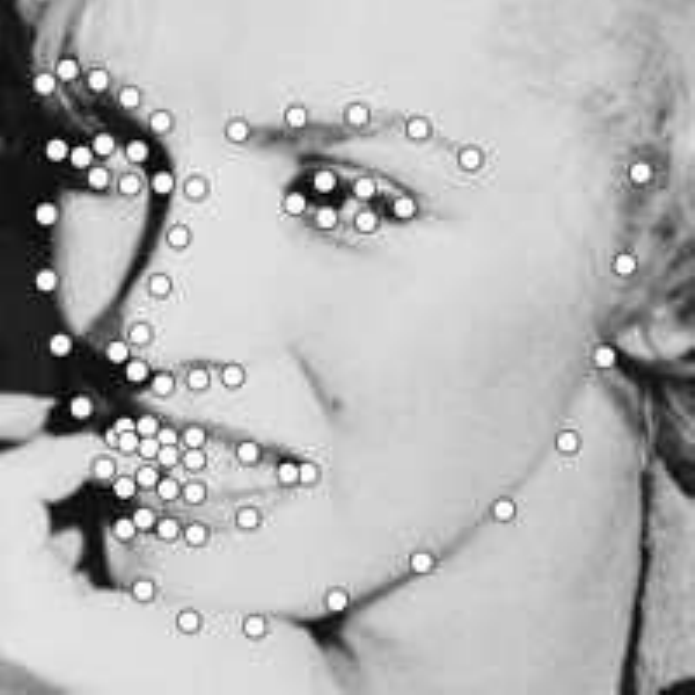}

\includegraphics[scale=0.22]{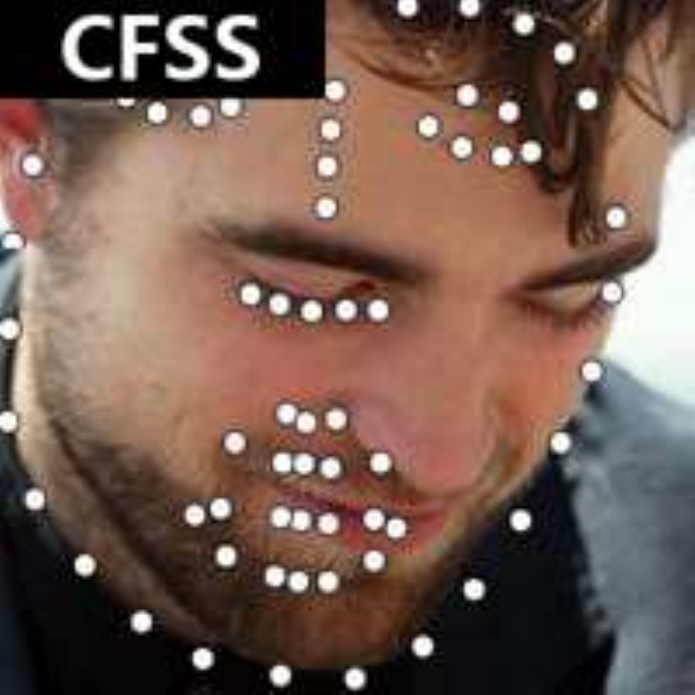}
\includegraphics[scale=0.22]{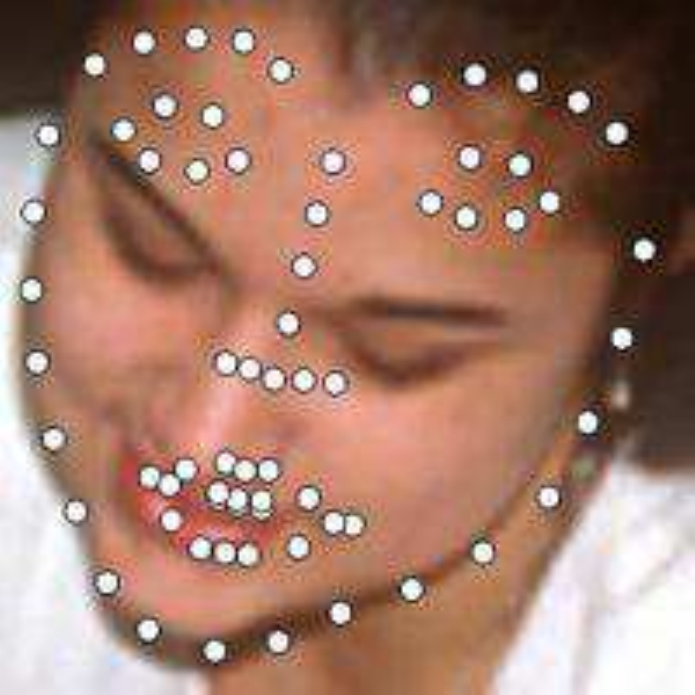}
\includegraphics[scale=0.22]{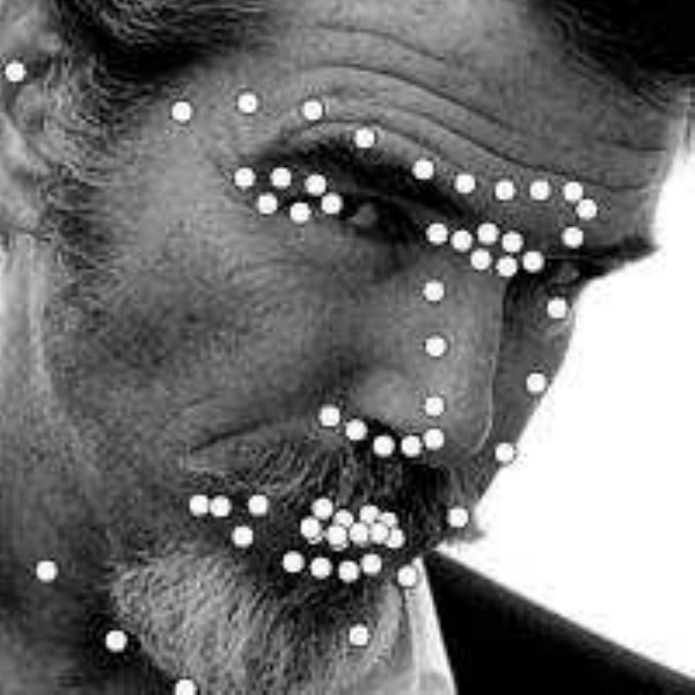}
\includegraphics[scale=0.22]{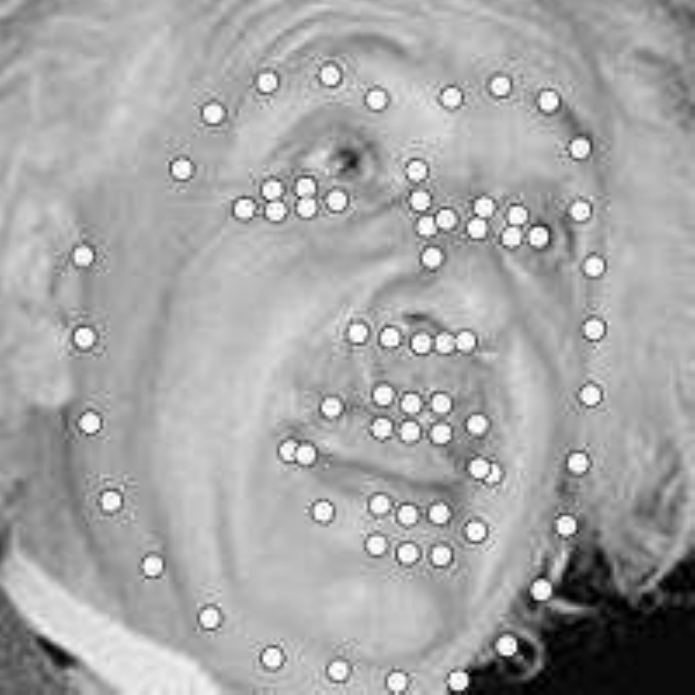}
\includegraphics[scale=0.22]{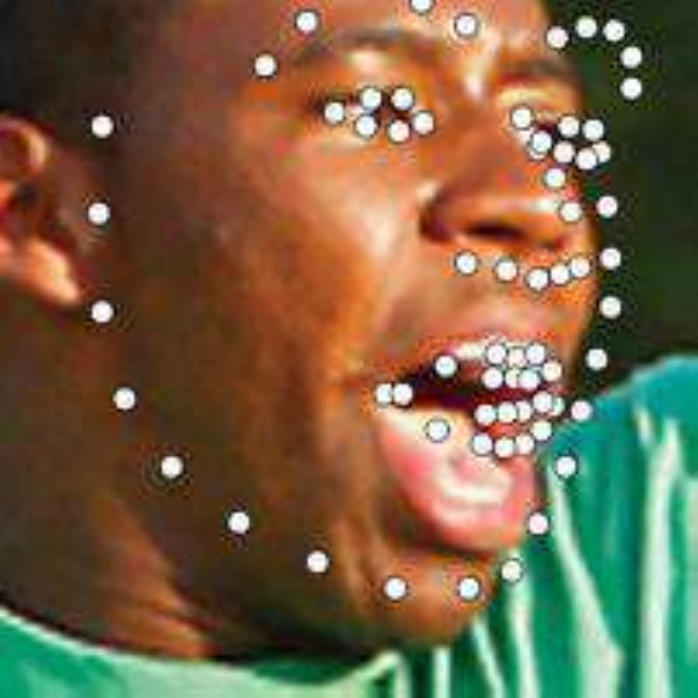}
\includegraphics[scale=0.22]{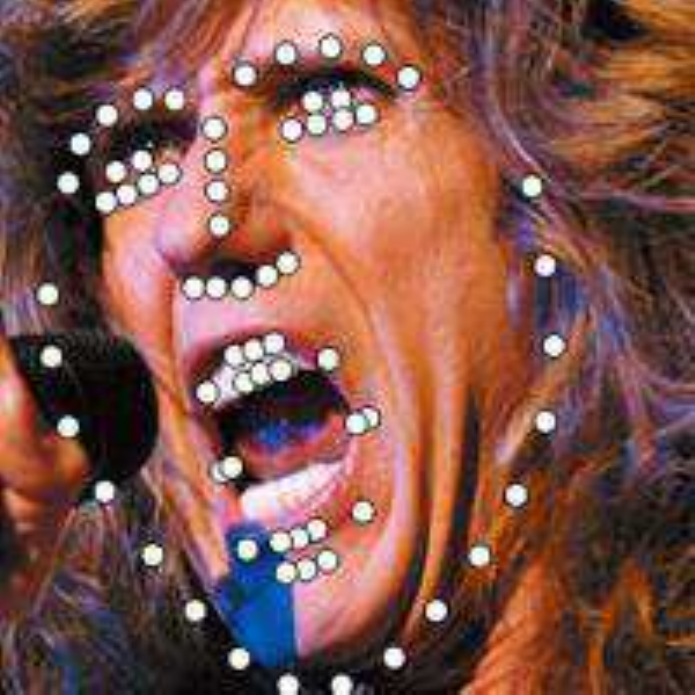}
\includegraphics[scale=0.22]{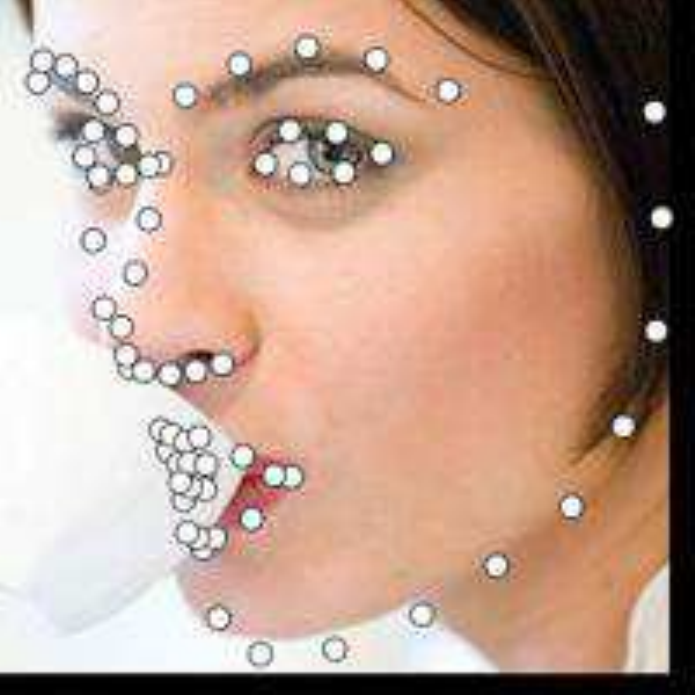}
\includegraphics[scale=0.22]{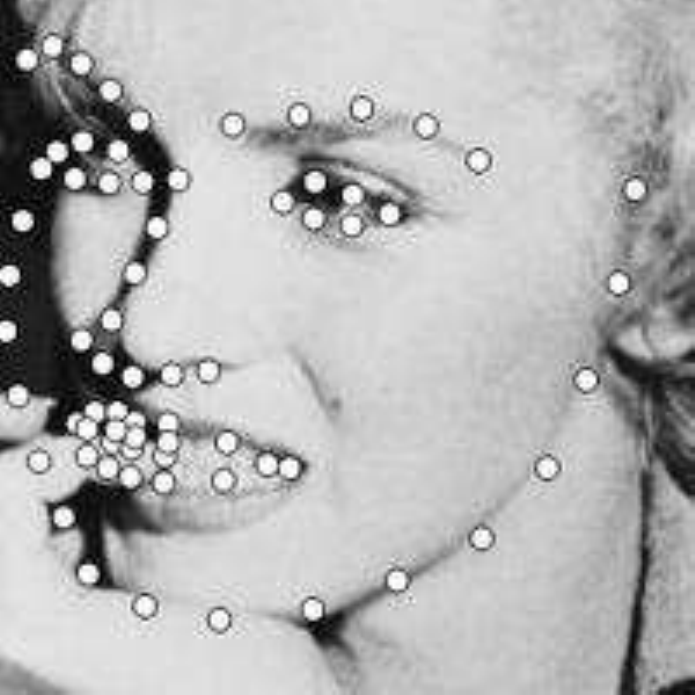}

\includegraphics[scale=0.22]{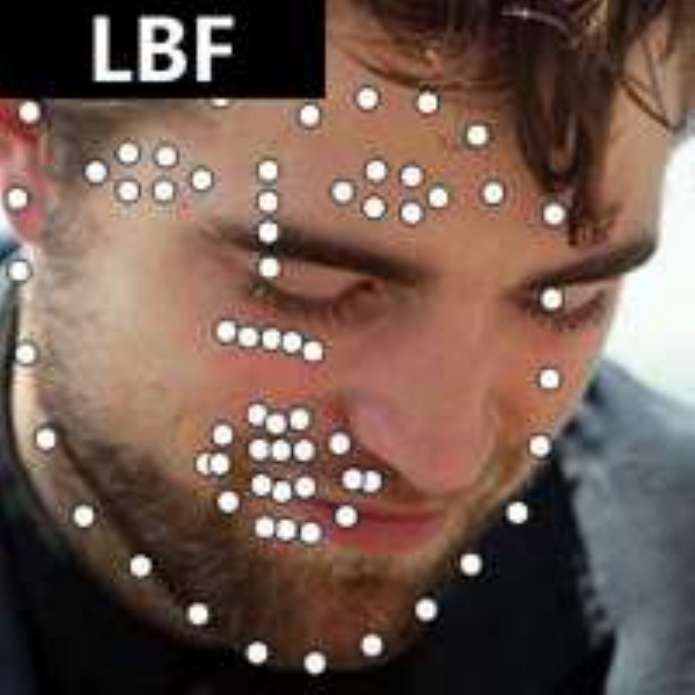}
\includegraphics[scale=0.22]{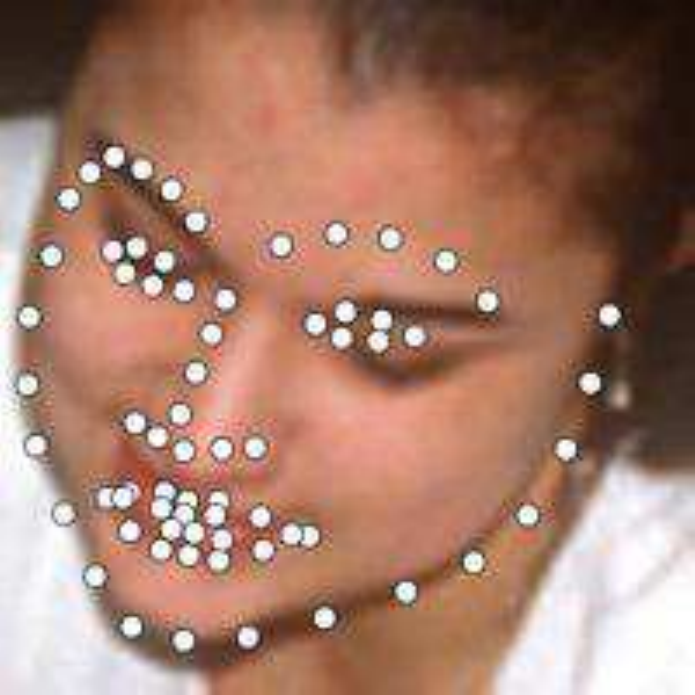}
\includegraphics[scale=0.22]{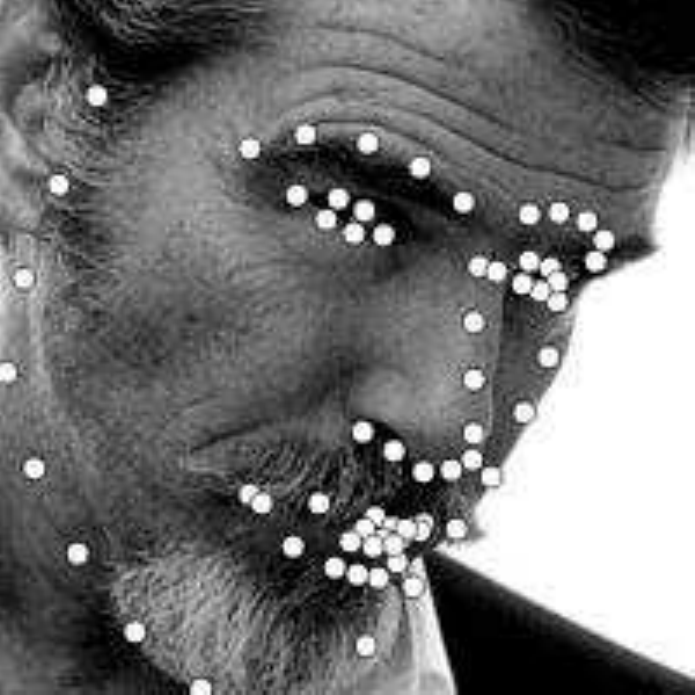}
\includegraphics[scale=0.22]{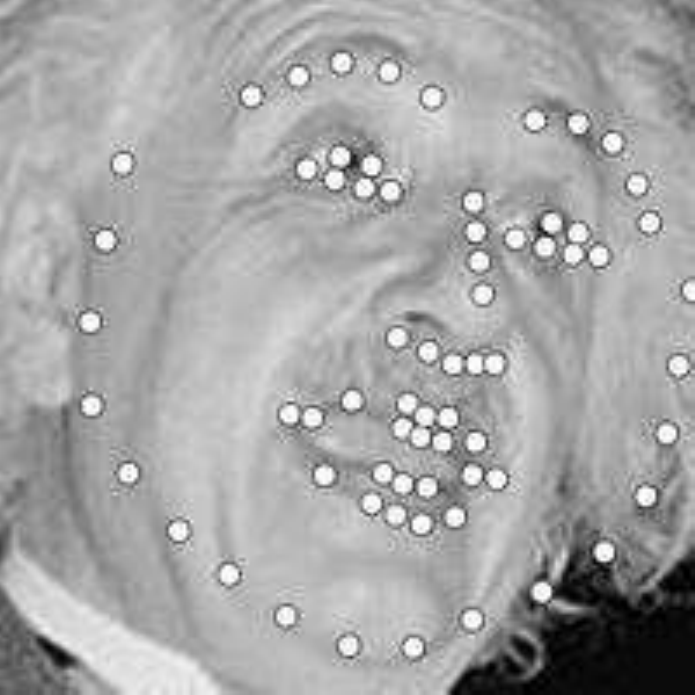}
\includegraphics[scale=0.22]{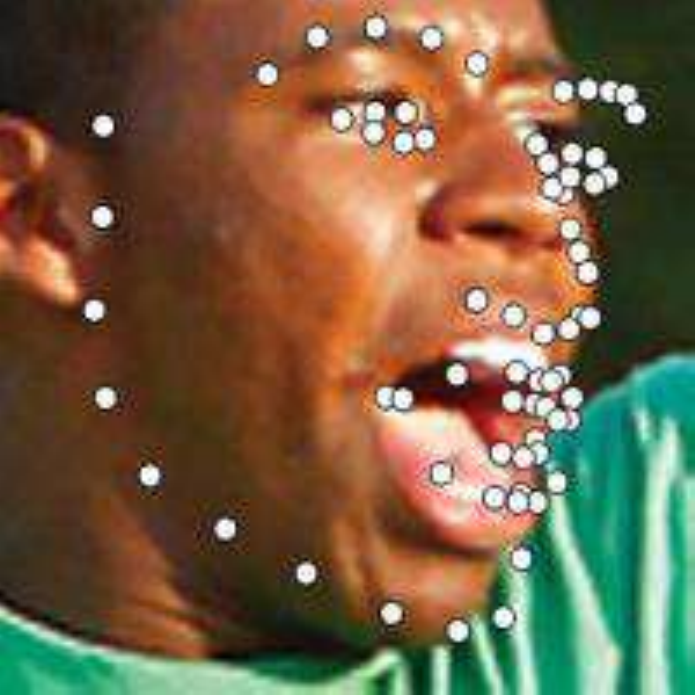}
\includegraphics[scale=0.22]{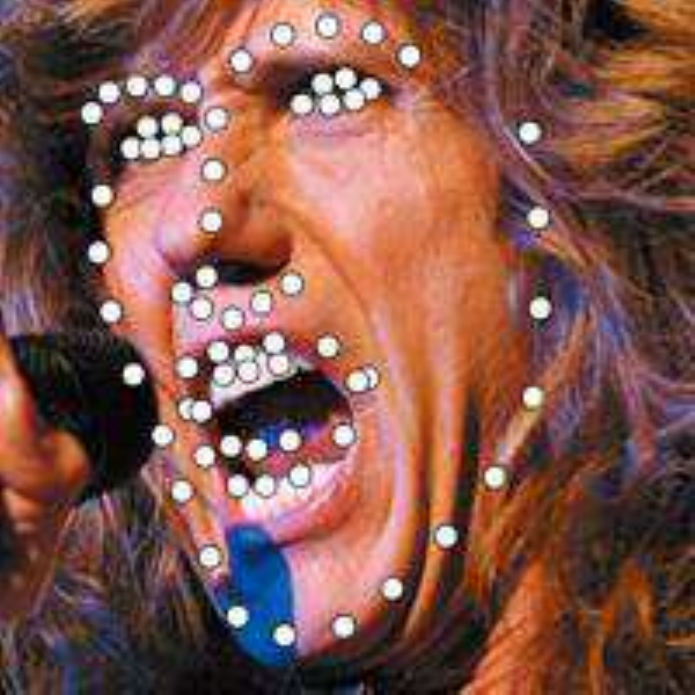}
\includegraphics[scale=0.22]{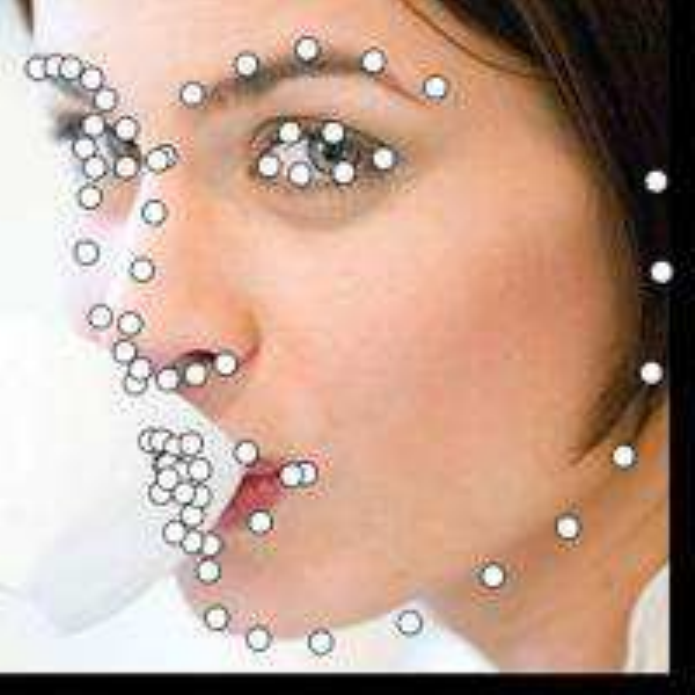}
\includegraphics[scale=0.22]{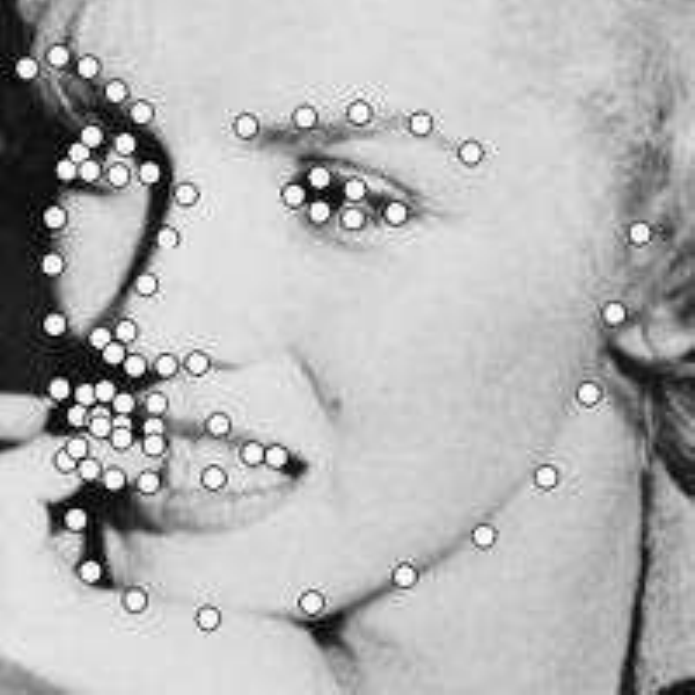}
\caption{Examples of LBF, CFSS, and MCL on challenging IBUG.}
\label{fig:compare IBUG}
\end{figure*}

\begin{figure*}[!htb]
  \centering
  \subfigure[CED for AFLW.]{
    \label{fig:CED:a} 
    \includegraphics[width=1.8in]{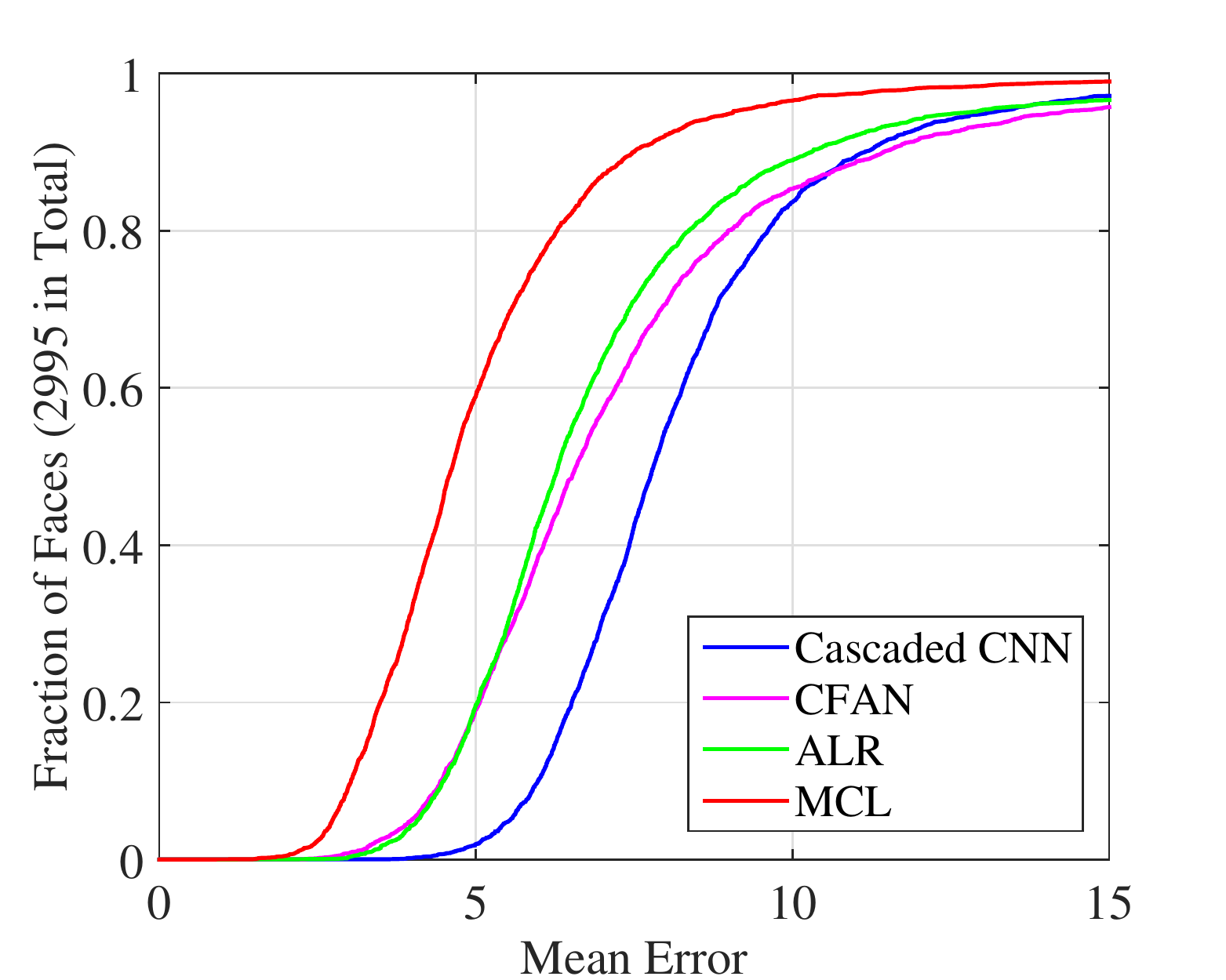}}
  \hspace{0in}
  \subfigure[CED for COFW.]{
    \label{fig:CED:b} 
    \includegraphics[width=1.8in]{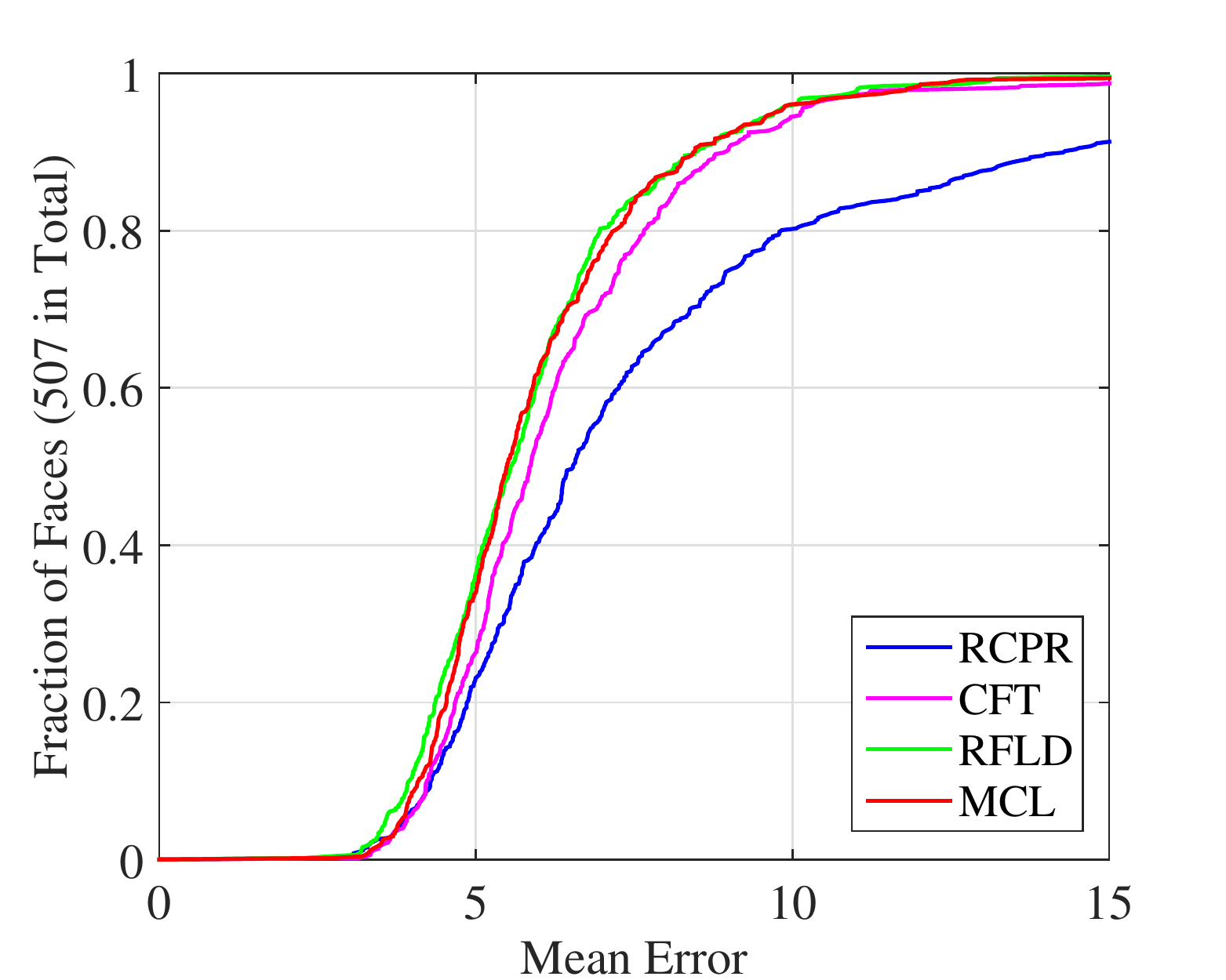}}
  \hspace{0in}
  \subfigure[CED for IBUG.]{
    \label{fig:CED:c} 
    \includegraphics[width=1.8in]{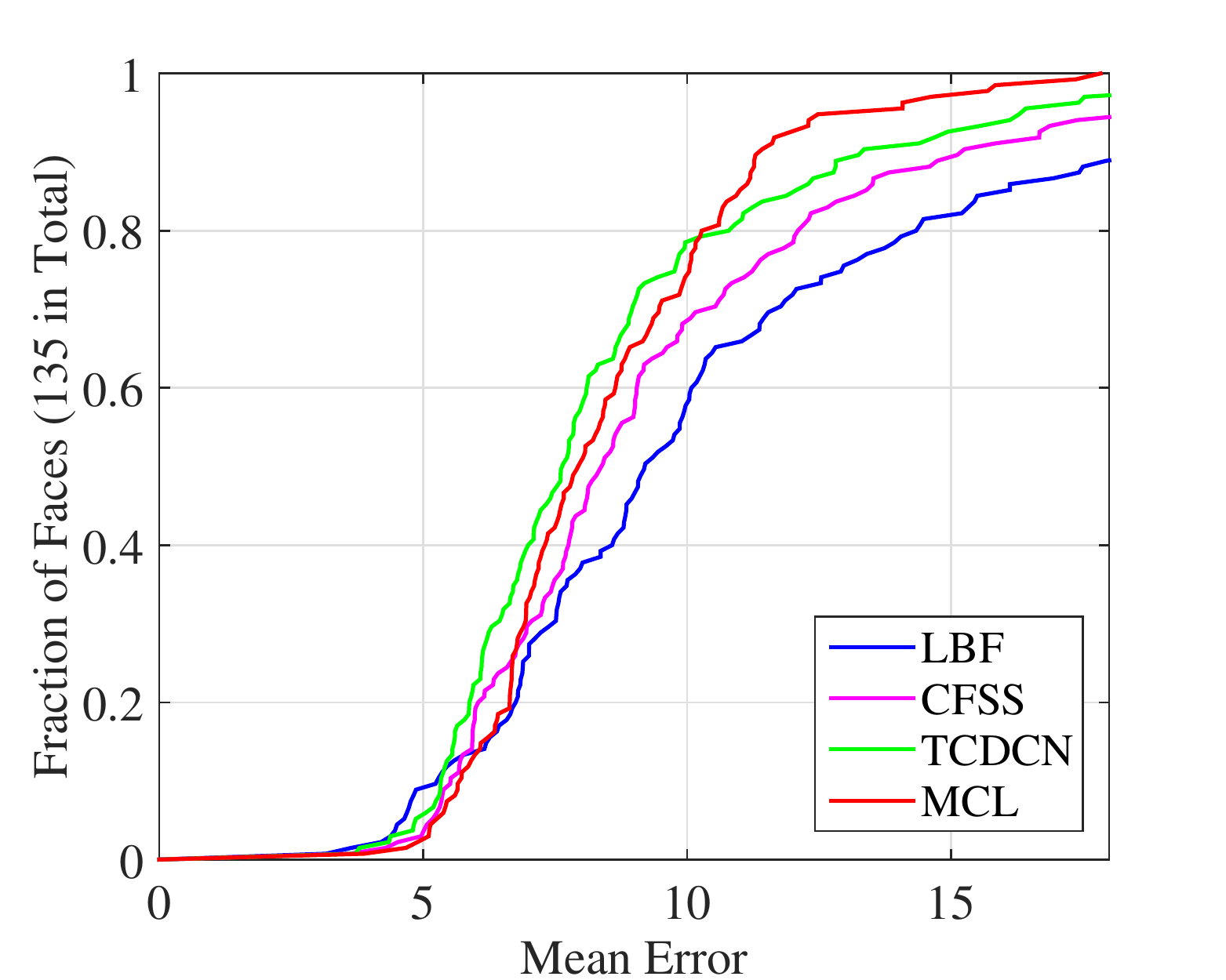}}
  \caption{Comparison of CED curves with previous methods on three benchmarks.}
  \label{fig:CED} 
\end{figure*}

We compare with other methods on several challenging images from AFLW and COFW respectively in Fig. \ref{fig:compare results}. Our method MCL indicates higher accuracy in the details than previous works. More examples on challenging IBUG are presented in Fig. \ref{fig:compare IBUG}. MCL demonstrates a superior capability of handling severe occlusions and complex variations of pose, expression, illumination. The CED curves of MCL and several state-of-the-art methods are shown in Fig. \ref{fig:CED}. It is observed that MCL achieves competitive performance on all three benchmarks.

\begin{table}[!htb]
\centering\caption{Average running speed of deep learning methods. The time of the face detection is excluded.}
\label{tab:comp_time}
\begin{tabular}{|*{3}{c|}}
\hline
Method & Speed (FPS) & Platform\\
\hline
CFAN \cite{zhang2014coarse} &43 &i7-3770 3.4 GHz CPU\\
TCDCN \cite{zhang2015learning} &50 &i5-6200U 2.3GHz CPU\\
CFT \cite{shao2016learning} &31 &i5-6200U 2.3GHz CPU\\
RAR \cite{xiao2016robust} &4 &Titan-Z GPU\\
\textbf{MCL} &\textbf{57} &i5-6200U 2.3GHz CPU\\
\hline
\end{tabular}
\end{table}

The average running speed of deep learning methods for detecting $68$ facial landmarks are presented in Table \ref{tab:comp_time}. Except for the methods tested on the i5-6200U 2.3GHz CPU, other methods are reported with the results in the original papers. Since CFAN utilizes multiple networks, it costs more running time. RAR achieves only $4$ FPS on a Titan-Z GPU, which cannot be applied to practical scenarios. Both TCDCN and our method MCL are based on only one network, so they show higher speed. Our method only takes $17.5$ ms per face on a single core i5-6200U 2.3GHz CPU. This profits from low model complexity and computational costs of our network. It can be concluded that our method is able to be extended to real-time facial landmark tracking in unconstrained environments.

\begin{table}[!htb]
\centering\caption{Results of mean error of pre-BM and BM on three benchmarks.}
\label{tab:val_GAP}
\begin{tabular}{|*{4}{c|}}
\hline
\multirow{2}*{Method} &AFLW &COFW &IBUG \\
&5 landmarks&29 landmarks&68 landmarks\\
\hline
pre-BM \cite{shao2017learning}&\textbf{5.61}&6.40&9.23\\
\textbf{BM}&5.67&\textbf{6.25}&\textbf{8.89}\\
\hline
\end{tabular}
\end{table}

\subsection{Ablation Study}

\subsubsection{Global Average Pooling vs. Full Connection}

Based on the previous version of our work \cite{shao2017learning}, the last max-pooling layer and the $D$-dimensional fully-connected layer are replaced with a convolutional layer and a Global Average Pooling layer\cite{lin2013network}. The results of the mean error of BM and the previous version (pre-BM) \cite{shao2017learning} are shown in Table \ref{tab:val_GAP}. It can be seen that BM performs better on IBUG and COFW but worse on AFLW than pre-BM. It demonstrates that Global Average Pooling is more advantageous for more complex problems with more facial landmarks. There are higher requirements for learned features when localizing more facial landmarks. For simple problems especially for localizing $5$ landmarks of AFLW, a plain network with full connection is more prone to being trained.

The difference between pre-BM and BM is the structure of learning the feature $\mathbf{x}$. The number of parameters for this part of pre-BM and BM are $(4\times4\times128+1)D=2,049D$ and $(3\times3\times128+1)D+2D+2D=1,157D$ respectively, where the three terms for BM correspond to the convolution, the expectation and variance of BN \cite{ioffe2015batch}, and the scaling and shifting of BN. Therefore, BM has a stronger feature learning ability with fewer parameters than pre-BM.

\begin{figure}[!htb]
  \centering
  \includegraphics[width=3.6in]{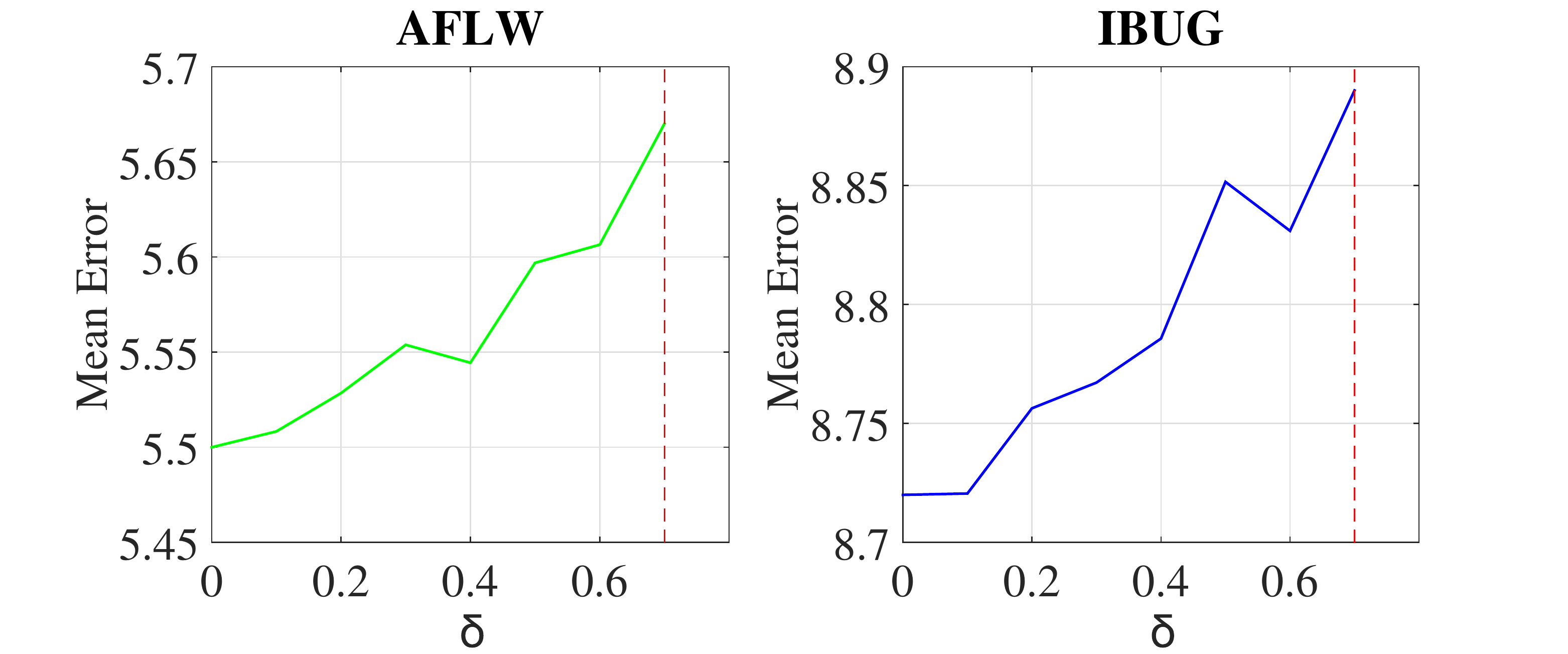}
  \caption{Mean error of WM on AFLW and IBUG with different $\delta$.}
  \label{fig:weight_perturb} 
\end{figure}

\subsubsection{Robustness of Weighting}

To verify the robustness of weighting, random perturbations are added to the weights of landmarks. In particular, we plus a perturbation $\delta$ to the weight of each of random $\lfloor n/2 \rfloor$ landmarks and minus $\delta$ to the weight of each of remaining $n-\lfloor n/2 \rfloor$ landmarks, where $\lfloor \cdot \rfloor$ refers to rounding down to the nearest integer. Fig. \ref{fig:weight_perturb} shows the variations of mean error of WM with the increase of $\delta$. When $\delta$ is $0.4$, WM can still achieves good performance. Therefore, weighting the loss of each landmark is robust to random perturbations. Even if different weights are obtained, the results will not be affected as long as the relative sizes of weights are identical.

\begin{table}[!htb]
\centering\caption{Results of mean error of landmarks of each cluster on IBUG.}
\label{tab:central_model}
\begin{tabular}{|*{4}{c|}}
\hline
Cluster&WM &Left Eye Model &Right Eye Model\\
\hline
Left Eye&8.09&\textbf{7.92}&8.10\\
Right Eye&7.73&7.55&\textbf{7.30}\\
Nose&6.19&6.42&6.59\\
Mouth&6.92&6.80&7.08\\
Left Contour&12.66&12.83&12.74\\
Chin&13.55&13.50&13.45\\
Right Contour&13.38&13.47&13.45\\
\hline
\end{tabular}
\end{table}

\subsubsection{Analysis of Shape Prediction Layers}

Our method learns each shape prediction layer respectively with a certain cluster of landmarks being emphasized. The results of WM and two shape prediction layers with respect to the left eye and the right eye on IBUG benchmark are shown in Table \ref{tab:central_model}. Compared to WM, the left eye model and the right eye model both reduce the alignment errors of their corresponding clusters. As a result, the assembled AM can improve the detection accuracy of landmarks of the left eye and the right eye on the basis of WM.

Note that the two models also improve the localization precision of other clusters. Taking the left eye model as an example, it additionally reduces the errors of landmarks of right eye, mouth, and chin, which is due to the correlations among different facial parts. Moreover, for the right eye cluster, the right eye model improves the accuracy more significantly than the left eye model. It can be concluded that each shape prediction layer emphasizes on the corresponding cluster respectively.

\begin{table}[!htb]
\centering\caption{Results of mean error of Simplified AM, AM, and Weighting Simplified AM. $29$ landmarks of COFW and $68$ landmarks of IBUG are evaluated.}
\label{tab:val_WS}
\begin{tabular}{|*{5}{c|}}
\hline
Method& \tabincell{c}{Weighting\\Fine-Tuning} &\tabincell{c}{Multi-Center\\Fine-Tuning} &COFW &IBUG\\
\hline
Simplified AM & &$\surd$ &6.08 &8.67\\
\textbf{AM} &$\surd$ &$\surd$ &\textbf{6.00} &\textbf{8.51}\\
\tabincell{c}{Weighting\\ Simplified AM} &$\surd$ &$\surd$ &6.05 &8.67\\
\hline
\end{tabular}
\end{table}

\subsubsection{Integration of Weighting Fine-Tuning and Multi-Center Fine-Tuning}

Here we validate the effectiveness of weighting fine-tuning by removing the weighting fine-tuning stage to learn a \emph{Simplified AM} from BM. Table \ref{tab:val_WS} presents the results of mean error of Simplified AM and AM respectively on COFW and IBUG. Note that Simplified AM has already acquired good results, which verifies the effectiveness of the multi-center fine-tuning stage. The accuracy of AM is superior to that of Simplified AM especially on challenging IBUG, which is attributed to the integration of two stages. A \emph{Weighting Simplified AM} from Simplified AM using the weighting fine-tuning stage is also learned, whose results are shown in Table \ref{tab:val_WS}. It can be seen that Weighting Simplified AM improves slightly on COFW but fails to search a better solution on IBUG. Therefore, we choose to use the multi-center fine-tuning stage after the weighting fine-tuning stage.

\begin{table}[!htb]
\centering\caption{Mean error (Error) and failure rate (Failure) of BM, WM, and AM. $5$ landmarks of AFLW, $29$ landmarks of COFW, and $68$ landmarks of IBUG are evaluated.}
\label{tab:allModel}
\begin{tabular}{|*{7}{c|}}
\hline
\multirow{2}*{Method}&\multicolumn{2}{c|}{AFLW}&\multicolumn{2}{c|}{COFW}&\multicolumn{2}{c|}{IBUG}\\
\cline{2-7}&Error&Failure&Error&Failure&Error&Failure\\
\hline
BM&5.67&4.43&6.25&5.13&8.89&27.43\\
WM&5.50&3.84&6.11&4.54&8.72&26.80\\
\textbf{AM}&\textbf{5.38}&\textbf{3.47}&\textbf{6.00}&\textbf{3.94}&\textbf{8.51}&\textbf{25.93}\\
\hline
\end{tabular}
\end{table}

\subsubsection{Discussion of All Stages}

\begin{figure}[!htb]
  \centering
  \includegraphics[width=3.2in]{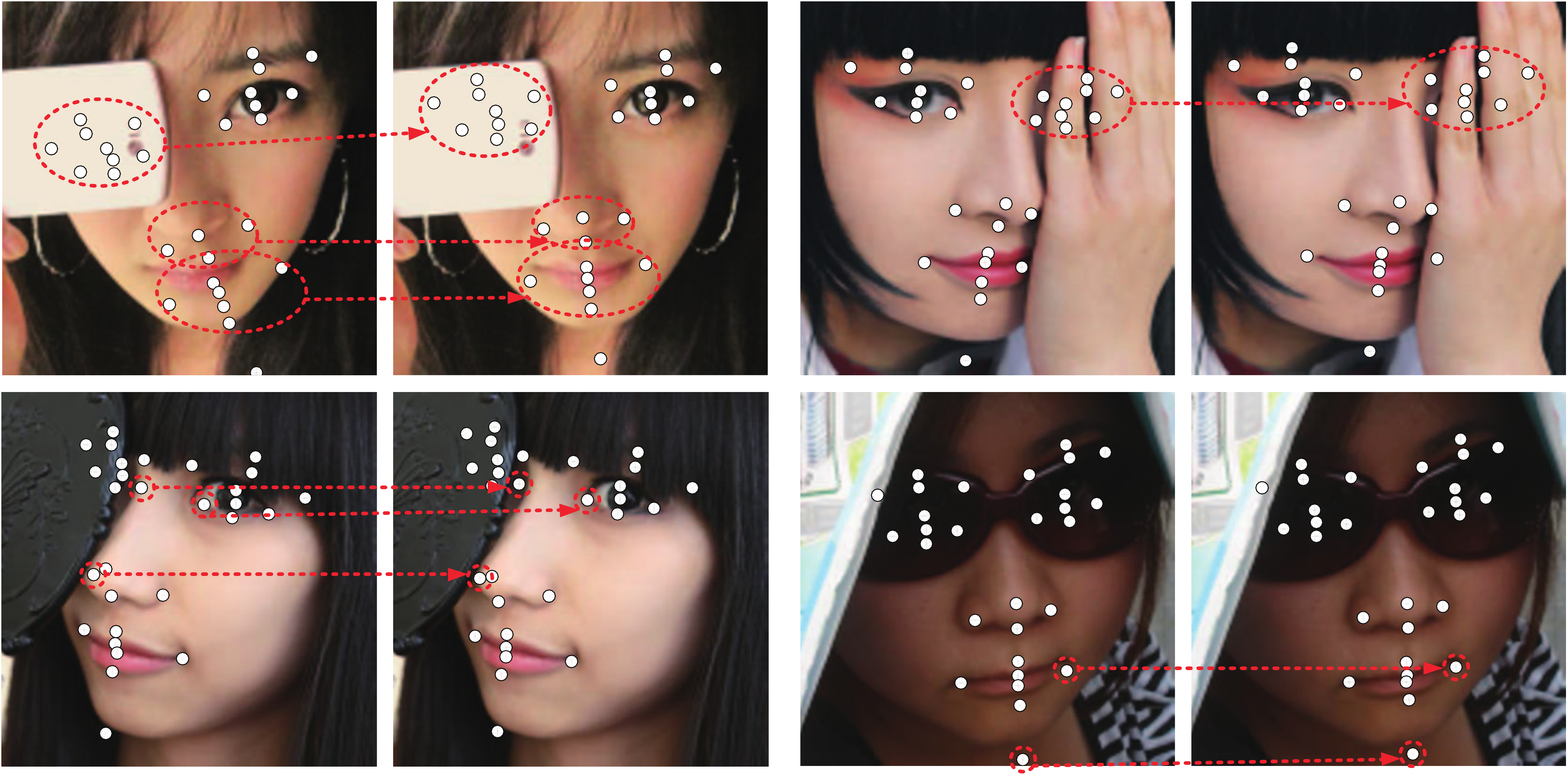}
  \caption{Examples of improvement for different facial landmarks from WM to AM on COFW dataset.}
  \label{fig:combine_validate_examples} 
\end{figure}

Table \ref{tab:allModel} summarizes the results of mean error and failure rate of BM, WM, and AM. It can be observed that AM has higher accuracy and stronger robustness than BM and WM. Fig. \ref{fig:combine_validate_examples} depicts the enhancement from WM to AM for several examples of COFW. The localization accuracy of facial landmarks from each cluster is improved in the details. It is because each shape prediction layer increases the detection precision of corresponding cluster respectively.

\subsection{MCL for Partially Occluded Faces}

\begin{figure}[!htb]
\centering
\includegraphics[scale=0.25]{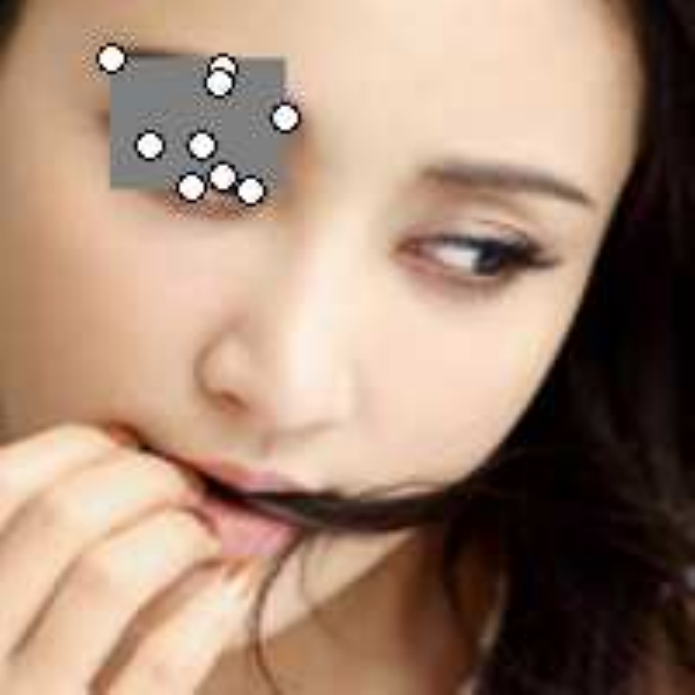}
\includegraphics[scale=0.25]{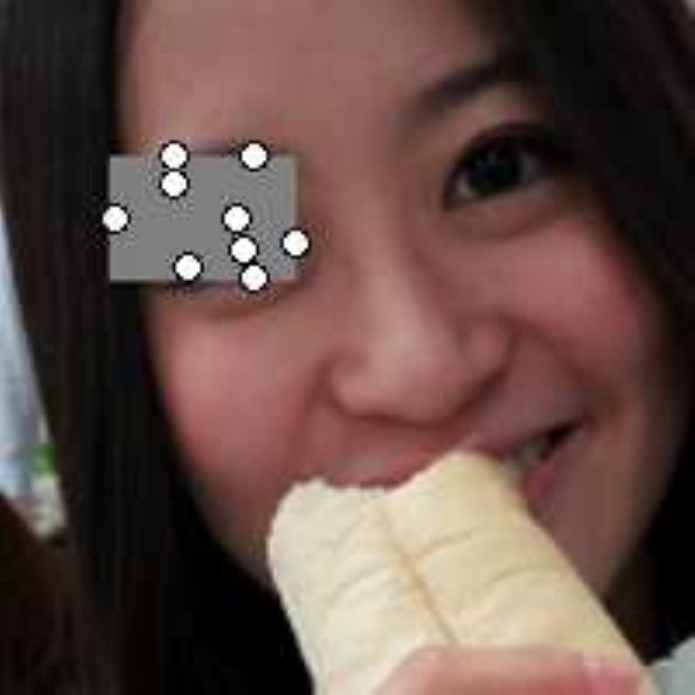}
\includegraphics[scale=0.25]{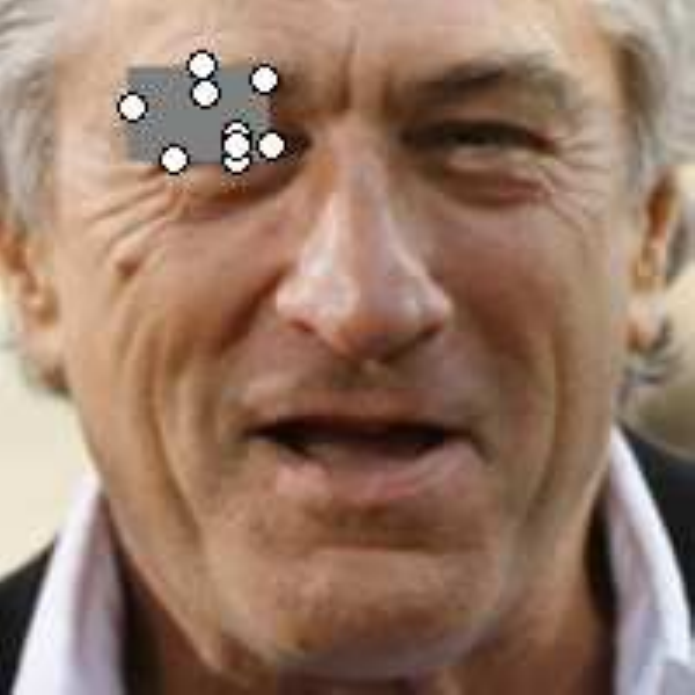}
\includegraphics[scale=0.25]{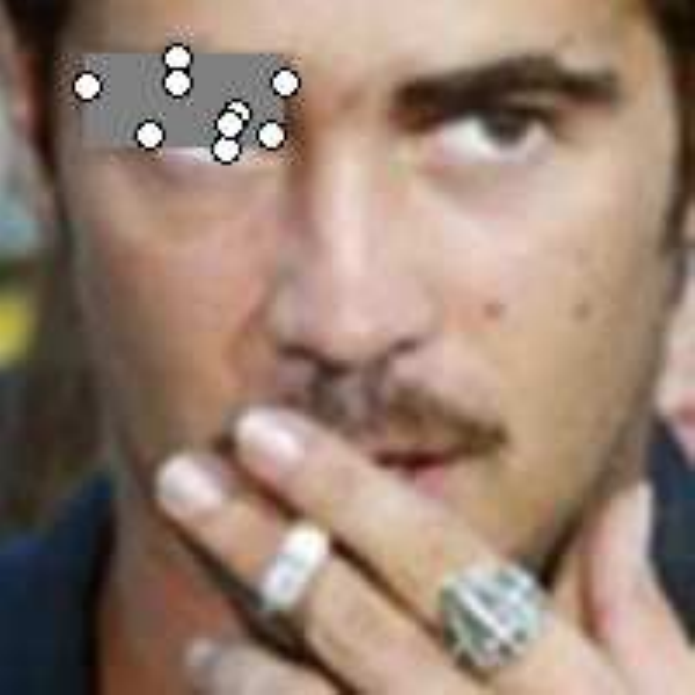}
\caption{Example faces from COFW with left eyes occluded.}
\label{fig:occlu_examples}
\end{figure}

The correlations among different facial parts are very useful for face alignment especially for partially occluded faces. To investigate the influence of occlusions, we directly use trained WM and AM without any additional processing for partially occluded faces. Randomly $30\%$ testing faces from COFW are processed with left eyes being occluded, where the tight bounding box covering landmarks of left eye is filled with gray color, as shown in Fig. \ref{fig:occlu_examples}.

Table \ref{tab:occlu_res} shows the mean error results for the left eye cluster and other clusters of WM and AM on COFW benchmark, where ``with (w/) occlusion (occlu.)'' denotes that left eyes of the testing faces are processed with handcrafted occlusions as illustrated in Fig. \ref{fig:occlu_examples}, and ``without (w/o) occlu.'' denotes that the testing faces are kept unchanged. Note that our method does not process occlusions explicitly, in which the training data is not performed handcrafted occlusions. After processing testing faces with occlusions, the mean error results of both WM and AM increase. Besides the results of landmarks from the left eye cluster, the results of remaining landmarks from other clusters become worse slightly. This is because different facial parts have correlations and the occlusions of the left eye influences results of other facial parts. Note that WM and AM still perform well on occluded left eyes with the mean error of $6.60$ and $6.50$ respectively, due to the following reasons. First, WM weights each landmark proportional to its alignment error, which exploits correlations among landmarks. Second, AM uses an independent shape prediction layer focusing on a certain cluster of landmarks with small weights $u_j>0$, $j\in Q^{i}$ in Eq. \ref{eq:landmark_am} for remaining landmarks, respectively, where correlations among landmarks are further exploited.

\begin{table}[!htb]
\centering\caption{Mean error results of WM and AM on COFW w/o occlu. and w/ occlu. respectively. Mean error of landmarks from the left eye cluster, and mean error of remaining landmarks from other clusters are both shown.}
\label{tab:occlu_res}
\begin{tabular}{|*{5}{c|}}
\hline
\multirow{2}*{Method} &\multicolumn{2}{c|}{Left Eye} &\multicolumn{2}{c|}{Others}\\
\cline{2-5}&w/o occlu.&w/ occlu.&w/o occlu.&w/ occlu.\\
\hline
WM&5.98&6.60&6.17&6.30\\
AM&5.88&6.50&6.06&6.18\\
\hline
\end{tabular}
\end{table}

\subsection{Weighting Fine-Tuning for State-of-the-Art Frameworks}

Most recently, there are a few well-designed and well-trained deep learning frameworks advancing the performance of face alignment, in which DAN \cite{kowalski2017deep} is a typical work. DAN uses cascaded deep neural networks to refine the localization accuracy of landmarks iteratively, where the entire face image and the landmark heatmap generated from the previous stage are used in each stage. To evaluate the effectiveness of our method extended to state-of-the-art frameworks, we conduct experiments with our proposed weighting fine-tuning being applied to DAN. In particular, each stage of DAN is first pre-trained and further weighting fine-tuned, where DAN with weighting fine-tuning is named DAN-WM.

\begin{table}[!htb]
\centering\caption{Results of mean error of DAN, re-DAN, and DAN-WM on IBUG.}
\label{tab:DAN_WM}
\begin{tabular}{|*{4}{c|}}
\hline
Method &DAN \cite{kowalski2017deep} &re-DAN &\textbf{DAN-WM}\\
\hline
IBUG &7.57 &7.97 &7.81\\
\hline
\end{tabular}
\end{table}

Note that the results of retrained DAN (re-DAN) using the published code\cite{kowalski2017deep} are slightly worse than reported results of DAN \cite{kowalski2017deep}. For a fair comparison, the results of mean error of DAN, re-DAN, and DAN-WM on IBUG benchmark are all shown in Table \ref{tab:DAN_WM}. It can be seen that the mean error of re-DAN is reduced from $7.97$ to $7.81$ after using our proposed weighting fine-tuning. Note that our method uses only a single neural network, which has a concise structure with low model complexity. Our network can be replaced with a more powerful one such as cascaded deep neural networks, which could further improve the performance of face alignment.

\section{Conclusion}
\label{sec:concl}

In this paper, we have developed a novel multi-center learning framework with multiple shape prediction layers for face alignment. The structure of multiple shape prediction layers is beneficial for reinforcing the learning process of each cluster of landmarks. In addition, we have proposed the model assembling method to integrate multiple shape prediction layers into one shape prediction layer so as to ensure a low model complexity. Extensive experiments have demonstrated the effectiveness of our method including handling complex occlusions and appearance variations. First, each component of our framework including Global Average Pooling, multiple shape prediction layers, weighting fine-tuning, and multi-center fine-tuning contributes to face alignment. Second, our proposed neural network and model assembling method allow real-time performance. Third, we have extended our method for detecting partially occluded faces and integrating with state-of-the-art frameworks, and have shown that our method exploits correlations among landmarks and can further improve the performance of state-of-the-art frameworks. The proposed framework is also promising to be applied for other face analysis tasks and multi-label problems.


%

%


\section*{Acknowledgments}

This work was supported by the National Natural Science Foundation of China (No. 61472245), and the Science and Technology Commission of Shanghai Municipality Program (No. 16511101300).

\ifCLASSOPTIONcaptionsoff
  \newpage
\fi



\bibliographystyle{IEEEtran}
\bibliography{IEEEabrv,references}
\end{document}